\documentclass[journal]{IEEEtran}
\ifCLASSINFOpdf
  \usepackage[pdftex]{graphicx}
\else
\fi
\usepackage{threeparttable}

\usepackage{multirow}
\usepackage{cmath}
\usepackage{algorithm, algorithmic}

\begin{document}
%
\title{Multi-Objective Matrix Normalization for Fine-grained Visual Recognition}
%
%
%

\author{Shaobo Min,
        Hantao Yao,~\IEEEmembership{Member,~IEEE},
        Hongtao Xie,
        Zheng-Jun Zha,
        and~Yongdong~Zhang,~\IEEEmembership{Senior~Member,~IEEE}

\thanks{This work is supported by the National Key Research and Development Program of China (2017YFC0820600), the National Nature Science Foundation of China (61525206, 61902399, 61771468, U1936210, 61622211), the National Postdoctoral Programme for Innovative Talents (BX20180358), the Youth Innovation Promotion Association Chinese Academy of Sciences (2017209), and the Fundamental Research Funds for the Central Universities under Grant WK2100100030.}

\thanks{S. Min, H. Xie, Z. Zha, and Y. Zhang are with the School of Information Science and Technology, University of Science and Technology of China, Hefei 230026, China. (e-mail: mbobo@mail.ustc.edu.cn; htxie@ustc.edu.cn; zhazj@ustc.edu.cn; zhyd73@ustc.edu.cn)}
\thanks{H. Yao is with the National Laboratory of Pattern Recognition, Institute of Automation, Chinese Academy of Sciences. Beijing, China. (e-mail: hantao.yao@nlpr.ia.ac.cn)}
\thanks{H. Xie and Y. Zhang are the corresponding authors.}%
}

%
%

\markboth{Journal of \LaTeX\ Class Files,~Vol.~14, No.~8, August~2015}%
{Shell \MakeLowercase{\textit{et al.}}: Bare Demo of IEEEtran.cls for IEEE Journals}
%



\maketitle

\begin{abstract}
Bilinear pooling achieves great success in fine-grained visual recognition (FGVC).
Recent methods have shown that the matrix power normalization can stabilize the second-order information in bilinear features, but some problems,~\emph{e.g.,} redundant information and over-fitting, remain to be resolved.
In this paper, we propose an efficient Multi-Objective Matrix Normalization (MOMN) method that can simultaneously normalize a bilinear representation in terms of square-root, low-rank, and sparsity.
These three regularizers can not only stabilize the second-order information, but also compact the bilinear features and promote model generalization.
In MOMN, a core challenge is how to jointly optimize three non-smooth regularizers of different convex properties.
To this end, MOMN first formulates them into an augmented Lagrange formula with approximated regularizer constraints.
Then, auxiliary variables are introduced to relax different constraints, which allow each regularizer to be solved alternately.
Finally, several updating strategies based on gradient descent are designed to obtain consistent convergence and efficient implementation.
Consequently, MOMN is implemented with only matrix multiplication, which is well-compatible with GPU acceleration, and the normalized bilinear features are stabilized and discriminative.
Experiments on five public benchmarks for FGVC demonstrate that the proposed MOMN is superior to existing normalization-based methods in terms of both accuracy and efficiency.
The code is available: https://github.com/mboboGO/MOMN.

\end{abstract}


%
\IEEEpeerreviewmaketitle

\section{Introduction}
Fine-grained visual categorization (FGVC) \cite{10.1007/978-3-319-10590-1_54,Lin2015a,yao2016coarse,He2017,ijcai2018-514,Wang2018} targets to distinguish objects with a subtle difference, which has attracted increasing attention recently. The large inner-class variances and subtle-class distinctions make FGVC a more challenging task than traditional image classification. To tackle this task, many related studies target to generate discriminative visual descriptions by exploring high-order image representations. 

\begin{figure}[t]
	\begin{center}
		\includegraphics[width=1\linewidth]{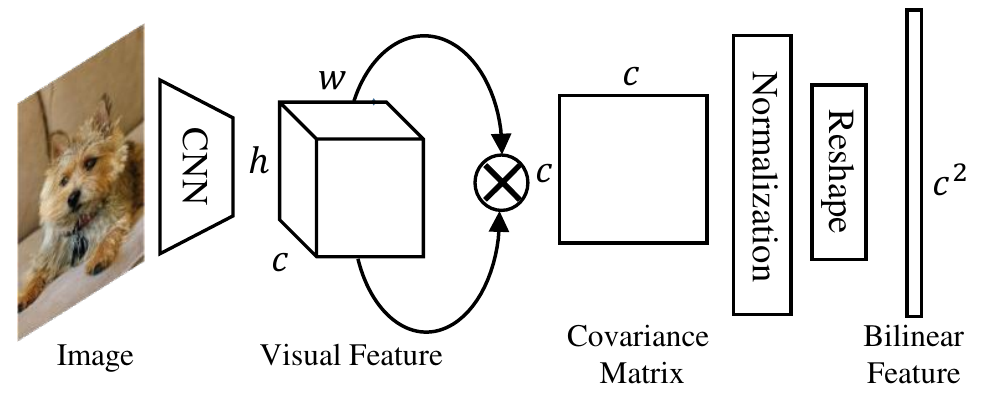}
	\end{center}
	\caption{A diagram of bilinear pooling for FGVC. The covariance matrix is obtained by first reshaping the visual feature $\boldsymbol{X}$ into $n\times c$ and then calculating $\boldsymbol{X}^{\top}\boldsymbol{X}$.}
	\label{fig:bp}
\end{figure}

Bilinear pooling is first introduced in \cite{Lin2015}, which pools the pairwise-correlated local descriptors into a global representation,~\emph{i.e.} bilinear feature, via outer-product operation.
A general diagram is shown in Figure~\ref{fig:bp}.
Compared with the global average pooling, bilinear pooling can model the second-order information of the input description, which has stronger representation and better discrimination.
However, the high-dimensional bilinear features~\emph{i.e.} about $250k$ in \cite{Lin2015}, have several defects: a) unstable second-order statistics,~\emph{e.g.,} visual burstiness problem \cite{sanchez2013image,wei2018grassmann,gou2018monet}; b) redundant features \cite{kong2017low}; c) over-fitting \cite{Lin2015}; and d) huge computational burden \cite{gao2016compact}.

\begin{table*}
\begin{center}
\caption{Comparison between MOMN and existing normalization-based methods for second-order representation. The good GPU support indicates that the method can be efficiently implemented on GPU platform. The good scalability indicates that the method can well accommodate multiple regularizers.} \label{tab:moti}
\begin{tabular}{l|c|c|c|c|c|c|c}
  \hline
  Frameworks&Normalization&Bottleneck&GPU support&Square-root&Low-rank&Sparsity&Scalability\\
  \hline
  \hline
  CBP~\cite{gao2016compact}&$l_2$&multiplication&good&$\times$&$\times$&$\times$&limited\\
  \hline
  MPN-COV~\cite{li2017second}&EIG&inverse&limited&$\surd$&$\surd$&$\times$&limited\\
  \hline
  G$^2$DeNet~\cite{Wang2017}&SVD&inverse&limited&$\surd$&$\surd$&$\times$&limited\\
  \hline
  Grass.Pool~\cite{wei2018grassmann}&SVD&inverse&limited&$\surd$&$\surd$&$\times$&limited\\
  \hline
  Impro. BCNN~\cite{lin2017improved}&NS Iter&multiplication&good&$\surd$&$\times$&$\times$&limited\\
  \hline
  iSQRT~\cite{Li2018}&NS Iter&multiplication&good&$\surd$&$\times$&$\times$&limited\\
  \hline
  MOMN (ours)&ADMM&multiplication&good&$\surd$&$\surd$&$\surd$&good\\
  \hline
\end{tabular}
\end{center}
\end{table*}

To address the above issues, compact bilinear pooling~\cite{gao2016compact}, as well as its variants \cite{cui2017kernel,cai2017higher}, approximates the out-product operation with efficient kernelized function.
Compared with original bilinear pooling, the kernelized bilinear pooling can significantly reduce the feature dimension without accuracy sacrifice.
However, the commonly used $l_2$ normalization,~\emph{i.e.,} element-wise square-root scaling, is too weak to suppress those ''burstiness" bilinear features \cite{wei2018grassmann,gou2018monet}.
To this end, matrix power and logarithm normalizations \cite{li2017second,Wang2017,lin2017improved,ionescu2015matrix,EnginWZL18} recently show great stabilization, by applying exponent or logarithm to the eigenvalues of bilinear features.
Despite the impressive performance, this kind of methods,~\emph{e.g.,} MPN-COV \cite{li2017second} and Grass. Pooling \cite{EnginWZL18}, depends heavily on the eigendecomposition (EIG) and singular value decomposition (SVD), which are not well-compatible with GPU platform.
Thus, they are limited to efficient deployment and application.
Recently, iSQRT \cite{Li2018} and Impro. BCNN \cite{lin2017improved} use the Newton-Schulz Iteration (NS Iter) \cite{Higham1997,Higham2009} to approximate matrix square-root normalization, which require only matrix multiplication.
However, the NS Iter is subjected to strict derivative conditions \cite{roberts1980linear}.
Thus, both iSQRT and Impro. BCNN cannot accommodate extra regularizers, such as low-rank and sparsity, which limits their further improvements in terms of compactness and generalization.
In summary, all the above methods cannot adapt to multi-objective normalization with efficient implementation for bilinear representation.
Table~\ref{tab:moti} gives a detailed comparison between different approaches.

In this paper, we propose a novel Multi-Objective Matrix Normalization (MOMN) algorithm, that can simultaneously normalize a bilinear representation in terms of square-root, low-rank, and sparsity. 
Besides the square-root regularizer for second-order stabilization, the low-rank regularizer can remove the trivial structure components, and the sparsity regularizer is used to suppress useless elements in the high dimensional features, which can alleviate the over-fitting problem.
However, these three regularizers are hard to jointly optimize due to non-smooth and different convex properties.

To resolve the above problem, MOMN first formulates an objective function with three target regularizers on the bilinear representation. 
Then, two auxiliary variables are introduced to loosen the correlations between different terms, which reformulate the objective function into a Lagrange formula.
According to the Alternating Direction Method of Multipliers (ADMM) \cite{boyd2011distributed} method, a closed-form solution can be obtained by alternately optimizing each variable in the reformulated function.
Further, several updating strategies based on gradient descent are designed to achieve consistent convergence and efficient implementation. 
For example, a covariance attention mechanism is developed to retain discriminative features during sparsity updating, and a coupled Newton iteration is adapted to MOMN for efficient convergence of square-root term.
Consequently, MOMN is implemented with only matrix multiplication, which is compatible with GPU platform and stabilized with back-propagation training.

Experiments show that, compared with previous methods, MOMN has several significant advantages: a) it has better generalization, which obtains the new state-of-the-art performance on five public benchmarks with five different backbones; b) the normalized bilinear representation is low-rank and sparse; c) it has fast training convergence and good compatibility with GPU acceleration; and d) MOMN can be easily extended to other matrix regularizers.
The overall contributions are summarized as following:
\begin{itemize}

\item We propose a Multi-Objective Matrix Normalization (MOMN) algorithm that can simultaneously regularize a second-order representation in terms of square-root, low-rank, and sparsity. 

\item Several updating strategies are developed that render MOMN to be optimized with fast and efficient convergence using only matrix multiplication.

\item We encapsulate MOMN into a plug-and-play layer, which can be easily extended to other useful regularizers and obtains the state-of-the-art performance on five public FGVC benchmarks.

\end{itemize}

The rest of this paper is organized as follows. Section~\ref{sec:rw} reviews the related works. Section~\ref{sec:method} illustrates the proposed MOMN. Section~\ref{sec:exp} provides the experiments on five benchmarks, and Section~\ref{sec:conclu} concludes the whole paper.

\section{Related Work}
\label{sec:rw}
The critical problem for fine-grained visual recognition \cite{gao2013learning,zhang2015detecting,liu2016hierarchical,yao2016coarse,He2017,Wang2018} is how to generate a robust and discriminative representation for fine-grained categories. 
As the artificial annotations, such as bounding boxes, attributes, and part annotations, are laborious to collect, a lot of representation learning methods~\cite{zhang2012attribute,zhang2014robust,He2017,yao2016coarse,gao2016compact,Simon2017,Wang2018,xu2019multi} have been proposed, which require only category labels to achieve accurate recognition.
In this section, we first talk about two mainstreams for FGVC, and then give a brief introduction to multi-objective optimization.
Finally, a discussion part summarizes the difference between the proposed Multi-Objective Matrix Normalization (MOMN) and related methods.

\subsection{Part-based Methods}
The part-based learning usually fuses the global and local representations by localizing discriminative part regions.
Besides employing the artificial annotations \cite{7226712,yao2016coarse,reed2016learning,yao2017dsp,Liu2017,Wei2018,zhang2019learning,ding2019selective}, many recent works can automatically learn the discriminative local regions with only image-level annotations.
For example, Jaderberg~\emph{et~al.}~\cite{Jaderberg2015} introduce a novel Spatial Transformer module to actively transform the feature maps into the localization networks.
Simon and Rodner~\cite{Simon2015} learn part models by finding constellations of neural activation patterns using neural networks.
Jonathan~\emph{et~al.}~\cite{krause2015fine} use a co-segmentation and alignment strategy to generate informative parts, and Fu~\emph{et~al.}~\cite{Fu2017} use attention modules to recursively localize interested regions and extract discriminative features in a mutually reinforced way. 
Recently, part-level and global-level localizations have proved that they can boost each other.
To localize discriminative object regions, He and Peng~\cite{He2017} propose to learn a whole-object detector through joint saliency extraction and co-segmentation.
Yao~\emph{et~al.}~\cite{yao2017autobd} design two complementary and elaborate part-level and object-level visual descriptions to capture robust and discriminative visual descriptions.
Moreover, MGE-CNN \cite{zhang2019learning} learns several expert networks to extract different region components, and S3N \cite{ding2019selective} selectively implements the semantic sampling on the informative regions to enhance feature representation.
Despite the strong representation by fusing global and local features, these methods require multi-stage training for part-based detectors, which complicates the training and inference processing.

\subsection{Second-order Pooling}
To generate discriminative image descriptors, the second-order statistics,~\emph{i.e.,} covariance matrices, have been explored in \cite{tuzel2006region,tuzel2007human,yao2016coarse,min2019adaptive}, which obtains impressive performance on human detection and texture classification.
Compared with original image representation, the second-order statistics show stronger discriminative power and better robustness, which are operated in the Riemannian manifolds.
Inspired by the second-order descriptors, the bilinear pooling is first proposed by Lin~\emph{et al.}~\cite{Lin2015} to capture the second-order information for FGVC. It applies the outer-product operation to local descriptors from two CNNs, followed by $l_2$ normalization,~\emph{i.e.,} element-wise square-root scaling.
The $l_2$ normalization is used to suppress some common features of high response,~\emph{i.e.,} visual burstiness problem \cite{sanchez2013image,wei2018grassmann}, thereby enhancing those specific and discriminative features relatively. 
Compared with part-based methods, bilinear pooling is a plug-and-play layer, which can be simply inserted in any backbone networks.
However, the bilinear representation suffers from several defects, such as the high dimension \cite{Lin2015} and over-fitting \cite{gao2016compact}.
 
To reduce the dimension of bilinear features, compact bilinear pooling~\cite{gao2016compact} utilizes the kernel function,~\emph{e.g.} Tensor Sketch \cite{charikar2004finding,Pham2013}, to approximate out-product operation.
Similarly, Yu~\emph{et al.}~\cite{yu2018hierarchical} use a dimension reduction projection before bilinear pooling to alleviate dimension explosion.
Compared with original bilinear pooling, these methods obtain comparable performance with much less computation burden.
Besides dimension reduction, many methods believe that kernelized bilinear pooling can capture higher-order information for stronger feature representation.
For example, Cui~\emph{et~al.}~\cite{cui2017kernel} introduce a kernel pooling method, that can capture arbitrarily order and non-linear features via compact feature mapping.
Cai~\emph{et~al.}~\cite{cai2017higher} propose a polynomial-kernel-based predictor to capture higher-order statistics of convolutional activations for modeling part interaction. 
However, all the above methods use the $l_2$ normalization, which is too weak to well stabilize high-order information,~\emph{i.e.,} suppressing ''burstiness" features, leading to slow convergence and low accuracy.

To this end, many methods~\cite{lin2017improved,li2017second,Li2018,EnginWZL18} explore the non-linearly scaling based on SVD and EIG to obtain better value stability for second-order representation.
Compared with raw elements in bilinear features, the singular vectors are more invariant to complex image contents, such as different object scales and illuminations, thereby having stronger feature representation.
For example, MPN-COV~\cite{li2017second} applies the power exponent to the eigenvalues of bilinear features, which surpass previous methods by a large margin.
Wang~\emph{et al.}\cite{Wang2017} further combine complementary first-order and second-order information via Gaussian embedding and matrix power normalization.
Recently, matrix square-root normalization, as a particular case of matrix power, has shown better numerically stability than matrix logarithm in~\cite{lin2017improved,gou2018monet,Li2018}. 
Therefore, Grass.Pool \cite{wei2018grassmann} and Gou~\emph{et al.}~\cite{gou2018monet} use a sub-matrix square-root layer before the bilinear pooling to obtain both high-order stabilization and low dimensional representation. 
Although the above methods achieve the state-of-the-art performance for FGVC, they depend heavily on SVD or EIG with matrix inverse operation, which is not well supported by GPU platform.
To this end, iSQRT~\cite{Li2018} and Impro. B-CNN \cite{lin2017improved} propose to use Newton-Schulz iteration to approximate matrix square-root normalization with only matrix multiplication, which obtain both efficient implementation and stabilized representation.
However, Newton-Schulz iteration is based on strict derivative conditions~\cite{roberts1980linear},~\emph{e.g.,} it should be continuously differentiable in a neighborhood of the root \cite{book} and the derivation should have coupled version to substitute the matrix inverse operation \cite{roberts1980linear}.
Thus, both iSQRT and Impro. BCNN cannot accommodate extra regularizers, such as low-rank and sparsity, to solve a multi-objective optimization problem with complex gradients.

Based on the above analysis, the proposed MOMN targets to normalize a bilinear representation, in terms of square-root, low-rank, and sparsity, simultaneously.
The extra low-rank constraint can eliminate the trivial components in bilinear features, which is beneficial to discriminative image representations.
Different from \cite{kong2017low} and \cite{Wang2017}, the low-rank constraints in MOMN is efficient and combined with other regularizations, without SVD or EIG.
Furthermore, the traditional image classification~\cite{Zhang2013,fan2018virtual} shows that imposing sparse constraint can significanly improve the feature generalization.
Unfortunately, these three non-smooth regularizers with different convex properties are hard to jointly optimize, which is the main challenge of this paper.

\begin{figure*}[t]
	\begin{center}
		\includegraphics[width=1\linewidth]{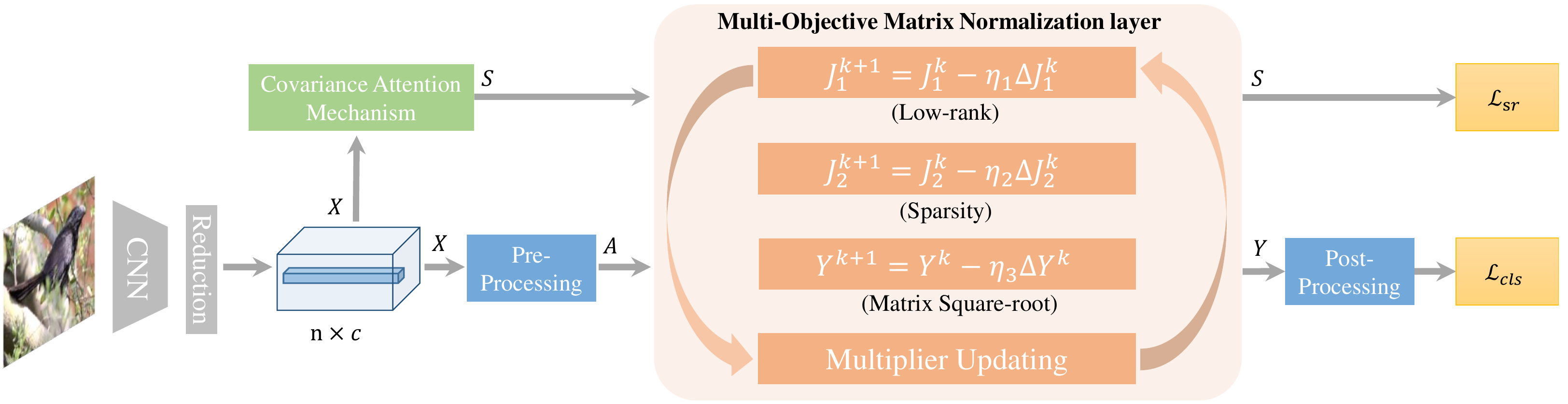}
	\end{center}
	\caption{The framework of MOMN. The Pre-Processing contains trace normalization, decentering, and covariance calculation. The Reduction layer consists of Conv., ReLU, and BN operations, which reduces the feature dimension. The four updating steps in MOMN are given in Sec.~\ref{sec:opt}. For simplifying, the classifier is omitted in $\mathcal{L}_{cls}$.}
	\label{fig:pl}
\end{figure*}

\subsection{Multi-objective Optimization}
Multi-objective optimization plays an important role in many tasks \cite{xu2017multi,wei2018robust,yang2018lr,sener2018multi}, which targets to optimize an objective function with several contrasting objectives.
A more detailed survey can be found in \cite{miettinen2012nonlinear,ehrgott2005multicriteria}.
In this paper, the most related work is about gradient-based multi-objective optimization \cite{fliege2000steepest,schaffler2002stochastic,desideri2012multiple,peitz2018gradient}, which is based on Karush-Kuhn-Tucker (KKT) conditions \cite{bertsekas1997nonlinear} and finds a descent direction for all objectives.
For example, the stochastic gradient \cite{you2014diverse,poirion2017descent} uses the properties of a common descent vector defined in the deterministic context to solve the multi-objective optimization.
In this paper, the gradient-based multi-objective optimization is extended to bilinear pooling framework under deep learning manner.
Different from existing multi-objective problems, additional requirements for MOMN are given that: a) the optimization process should be stabilized with deep model training, and b) the optimization process should be compatible with GPU platform.

\subsection{Discussion}
Finally, we discuss the main differences between MOMN and related FGVC methods.
First, some early methods,~\emph{e.g.,} S-SJE~\cite{reed2016learning} and PS-CNN~\cite{huang2016part}, use the additional annotations, ~\emph{e.g.,} category attributes, part regions, and key points, as auxiliary supervision to enhance feature representations.
However, the auxiliary annotations are usually hard to obtain, thus MOMN is designed that requires only image-level annotations.

Second, recent part-based methods, such as the Jonathan~\emph{et~al.}~\cite{krause2015fine} and MGE-CNN \cite{zhang2019learning}, leverage the strong attention mechanism to automatically localize the informative regions without part annotations.
Different from aggregating multiple part embeddings, MOMN focus on exploring the second-order visual statistic to enhance the feature representation.
Thus, MOMN is exactly complementary to the part-based methods.

Third, MOMN is based on the bilinear pooling method \cite{Lin2015}.
Specifically, iSQRT \cite{Li2018} and Impro. BCNN \cite{lin2017improved} replace the original $l_2$ norm with the matrix square-root normalization and develop an iterative version to further stabilize the second-order statistics, efficiently.
However, these methods, including other bilinear variants, cannot be extended to accommodate multiple important regularizers, such as the low-rank and sparsity.
Thus, our MOMN provides a much more generalized formulation that can simultaneously normalize the bilinear representation in terms of square-root, low-rank, and sparsity regularizers.
Besides, the low-rank BCNN \cite{kong2017low} utilizes the convolution layers to reduce the backbone  channel for a low-rank representation, which is much different from MOMN.
Instead, MOMN directly suppresses the trivial eigenvalues in the Riemannian manifold, which is much more effective.

In summary, the proposed MOMN is a new method that can accommodate multi-objective regularizers for bilinear representation.

\section{Multi-Objective Normalization}
\label{sec:method}
In this section, we first give the formulation of the Multi-Objective Matrix Normalization (MOMN), and then introduce the optimization details.
Finally, an implementation of MOMN is illustrated, of which the flowchart is shown in Figure~\ref{fig:pl}.

\subsection{Problem Formulation}
\label{sec:pf}
Given an input image, we define $\widehat{\boldsymbol{X}}\in R^{h\times w\times d}$ as the feature map of the last convolution layer, where $h,w$ and $d$ are the height, width, and channel.
To reduce the dimension of bilinear features, a dimension reduction layer,~\emph{i.e.,} conv+bn+relu, is employed, which generates $\boldsymbol{X}\in R^{h\times w\times c}$ and $c\ll d$.
For simplicity, $\boldsymbol{X}$ is reshaped to $n\times c$, where $n=h\times w$.
Based on reshaped $\boldsymbol{X}$, the bilinear feature,~\emph{i.e.}~covariance matrix, can be obtained by $\boldsymbol{A}=\boldsymbol{X}^{\top}\boldsymbol{C}\boldsymbol{X}$, where $\boldsymbol{C}=\frac{1}{n}(\boldsymbol{I}_{n\times n}-\frac{1}{n}\boldsymbol{1}_{n\times n})$, $\boldsymbol{I}_{n\times n}$ is the identity matrix, and $\boldsymbol{1}_{n\times n}$ is the matrix of all ones.
Notably, $\boldsymbol{A}$ is a symmetric positive definite (SPD) matrix, containing second-order information.
The target of MOMN is to normalize $\boldsymbol{A}$ in terms of: a) stabilizing second-order information; b) low-rank for redundancy reduction; and c) sparsity for good generalization.

By considering the above constraints, the objective function of MOMN is defined as Eq.~\eqref{eq:obj_f}:
\begin{eqnarray}
\label{eq:obj_f}
\min_{\boldsymbol{Y}}||\boldsymbol{Y}^{2}-\boldsymbol{A}||^2_{F}+\beta_1||\boldsymbol{Y}||_{*}+\beta_2||\boldsymbol{Y}||_{1},
\end{eqnarray}
where $\boldsymbol{Y}$ is the target regularized feature and the output of MOMN.
$||\cdot||_{F}$ is the Frobenius norm, which constrains $\boldsymbol{Y}$ to be the unique square root of $\boldsymbol{A}$.
$||\cdot||_{*}$ is the nuclear norm as the approximated rank function, which calculates the summation of singular values of an input matrix.
$||\cdot||_1$ is the $l_1$ norm to approximate $l_0$ norm for sparsity.
$\beta_1$ and $\beta_2$ are the hyper-parameters to balance the effects of different constraints.

The three terms in Eq.~\eqref{eq:obj_f} correspond to matrix regularizers of square-root, low-rank, and sparsity, respectively. Compared with related works \cite{li2017second,Li2018}, Eq.~\eqref{eq:obj_f} is a more general representation, \emph{e.g.,} matrix square-root normalization is a special case of $\beta_1=\beta_2=0$.
However, it is non-trivial to optimize Eq.~\eqref{eq:obj_f}, because: a) $||\cdot||_{*}$ and $||\cdot||_{1}$ are non-smooth; b) three constraint terms have different convex properties.
Therefore, we propose a Multi-Objective Matrix Normalization to optimize Eq.~\eqref{eq:obj_f}. In the following, we will give the detailed optimization process of MOMN.

\subsection{Optimization}
\label{sec:opt}
Since the joint optimization for three intractable terms in Eq.~\eqref{eq:obj_f} is infeasible, two auxiliary variables $\boldsymbol{J}_1$ and $\boldsymbol{J}_2$ are introduced to process them independently. 
As a consequence, the original objective function Eq.~\eqref{eq:obj_f} is reformulated as following equivalent formula: 
\begin{eqnarray}
\label{eq:obj_a}
\begin{split}
\min_{\boldsymbol{J}_1,\boldsymbol{J}_2,\boldsymbol{Y}}&||\boldsymbol{Y}^{2}-\boldsymbol{A}||^2_{F}+\beta_1||\boldsymbol{J}_1||_{*}+\beta_2||\boldsymbol{J}_2||_{1},\\
s.t.& ~~~~~~~ \boldsymbol{J}_1=\boldsymbol{Y},\boldsymbol{J}_2=\boldsymbol{Y}.
\end{split}
\end{eqnarray}

Compared with Eq.~\eqref{eq:obj_f}, Eq.~\eqref{eq:obj_a} is a constrained optimization problem with three relaxed constraint terms.
Thus, we further convert Eq.~\eqref{eq:obj_a} into an unconstrained form of Eq.~\eqref{eq:obj_l} by introducing augmented Lagrange multipliers.

\begin{eqnarray}
\label{eq:obj_l}
\begin{split}
\min_{\boldsymbol{J}_1,\boldsymbol{J}_2,\boldsymbol{Y}}&||\boldsymbol{Y}^{2}-\boldsymbol{A}||^2_{F}+\beta_1||\boldsymbol{J}_1||_{*}+\beta_2||\boldsymbol{J}_2||_{1}\\
&+tr(\boldsymbol{L}_1^{\top}(\boldsymbol{J}_1-\boldsymbol{Y}))+\frac{\mu_1}{2}||\boldsymbol{J}_1-\boldsymbol{Y}||^2_{F}\\
&+tr(\boldsymbol{L}_2^{\top}(\boldsymbol{J}_2-\boldsymbol{Y}))+\frac{\mu_2}{2}||\boldsymbol{J}_2-\boldsymbol{Y}||^2_{F},
\end{split}
\end{eqnarray}
where $\boldsymbol{L}_1$ and $\boldsymbol{L}_2$ are the Lagrange multipliers, $\mu_1$ and $\mu_2$ are positive penalty parameters, and $tr(\cdot)$ gets the matrix trace.
Instead of simple Lagrange multipliers with only linear penalty, we use the augmented Lagrange multipliers to convert the constrained Eq.~\eqref{eq:obj_a} into the unconstrained Eq.~\eqref{eq:obj_l} by introducing additional quadratic penalty terms, which can improve convergence speed and stabilization \cite{schaffler2002stochastic,miettinen2012nonlinear}. 
The reformlated objective function in Eq.~\eqref{eq:obj_l} can be solved under the Alternating Direction Method of Multipliers (ADMM) \cite{boyd2011distributed} framework.
Thus, a closed-form solution of Eq.~\eqref{eq:obj_l} can be obtained by alternately updating each variable.
In the following, we will illustrate how to update each term in Eq~\eqref{eq:obj_l}, with consistent convergence and efficient implementation.
The initialization of variables are: $\boldsymbol{J}_1^0=\boldsymbol{J}_2^0=\boldsymbol{Y}^0=\boldsymbol{A}$, and $\boldsymbol{L}_1^0=\boldsymbol{L}_2^0=\boldsymbol{0}$.
The integrated algorithm is shown in Algorithm~\ref{alg:MOMN}.

\textbf{Step 1: Update $\boldsymbol{J}_1$ while fixing other variables:} At the $k$-th iteration, we update $\boldsymbol{J}_1$ with fixed $\boldsymbol{J}_2$ and $\boldsymbol{Y}$. Therefore, the irrelevant terms in Eq.~\eqref{eq:obj_l} can be removed during updating $\boldsymbol{J}_1$:
\begin{equation}
\label{eq:obj_var_J}
\small
\boldsymbol{J}_1^{k+1}=\arg\min_{\boldsymbol{J}_1^{k}}\beta_1||\boldsymbol{J}_1^{k}||_{*}+\frac{\mu_1}{2}||\boldsymbol{J}^{k}_1-\boldsymbol{Y}^{k}+\frac{1}{\mu_1}\boldsymbol{L}_1^{k}||^2_{F}.
\end{equation}

A general solution for Eq~\eqref{eq:obj_var_J} is to use SVD or EIG, which is based on matrix inverse and not well-compatible with GPU acceleration \cite{Zhang2013,Wei2018,8499082}.
Instead, we employ the gradient descent algorithm to update $\boldsymbol{J}_1$ to make the algorithm GPU-friendly by:
\begin{eqnarray}
\label{eq:update_J1_v1}
\small
\begin{split}
\boldsymbol{J}^{k+1}_1=&\boldsymbol{J}^{k}_1-\eta_1\frac{\partial (\beta_1||\boldsymbol{J}_1^{k}||_{*}+\frac{\mu_1}{2}||\boldsymbol{J}^{k}_1-\boldsymbol{Y}^{k}+\frac{1}{\mu_1}\boldsymbol{L}_1^{k}||^2_{F})}{\partial\boldsymbol{J}_1^{k}}\\
=&\boldsymbol{J}^{k}_1-\eta_1[\beta_1\frac{\partial ||\boldsymbol{J}_1^{k}||_{*}}{\partial\boldsymbol{J}_1^{k}}+\mu_1(\boldsymbol{J}^k_1-\boldsymbol{Y}^k)+\boldsymbol{L}^k_1],
\end{split}
\end{eqnarray}
where $\eta_1$ is the step length.
Since $\boldsymbol{J}_1^k$ is a SPD matrix, its singular values and eigenvalues are equal. Thereby we have $
\frac{\partial ||\boldsymbol{J}_1^{k}||_{*}}{\partial\boldsymbol{J}_1^{k}}=\frac{\partial (tr(\boldsymbol{J}_1^{k}))}{\partial\boldsymbol{J}_1^{k}}=\boldsymbol{I}$.
Thus, Eq.~\eqref{eq:update_J1_v1} becomes:
\begin{eqnarray}
\label{eq:update_J_1_}
\boldsymbol{J}^{k+1}_1=&\boldsymbol{J}^{k}_1-\eta_1[\beta_1\boldsymbol{I}+\mu_1(\boldsymbol{J}^k_1-\boldsymbol{Y}^k+\frac{1}{\mu_1}\boldsymbol{L}^k_1)].
\end{eqnarray}

\begin{algorithm}
	\renewcommand{\algorithmicrequire}{\textbf{Input:}}
	\renewcommand{\algorithmicensure}{\textbf{Output:}}
	\caption{Multi-Objective Matrix Normalization}
	\label{alg:MOMN}
	\begin{algorithmic}[1]
		\REQUIRE Feature $\boldsymbol{X}$, hyper-parameters $\beta_1$, $\beta_2$, and optional covariance attention map $\boldsymbol{S}$.
		\STATE Calculating covariance matrix by $\boldsymbol{A}=\boldsymbol{X}^{\top}\boldsymbol{C}\boldsymbol{X}$.
		\STATE Pre-normalization by $\boldsymbol{A}\leftarrow\frac{1}{tr(\boldsymbol{A})}\boldsymbol{A}$.
		\STATE Initialize $\boldsymbol{Y}^0=\boldsymbol{J}_1^0=\boldsymbol{J}_2^0=\boldsymbol{A}$, $\boldsymbol{L}_1^0=\boldsymbol{L}_2^0=\boldsymbol{0}$, $\boldsymbol{Z}^0=\boldsymbol{I}$, $\mu_1>0$, $\mu_2>0$, $\rho>1$.
		\FOR{$k=1:K$}
		\STATE Update $\boldsymbol{J}_1^{k+1}$ as $\boldsymbol{J}^{k+1}_1=\boldsymbol{Y}^{k}-\frac{1}{\mu_1}\boldsymbol{L}_1^{k}-\frac{\beta_1}{\mu_1}\boldsymbol{I}$.
		\STATE Update $\boldsymbol{J}_2^{k+1}$ as ~$\boldsymbol{J}_2^{k+1}=\boldsymbol{Y}^{k}-\frac{1}{\mu_2}\boldsymbol{L}_{2}^{k}-\frac{\beta_2}{\mu_2}(1-\boldsymbol{S})\boldsymbol{J}_{2}^{k}$ or $\boldsymbol{Y}^{k}-\frac{1}{\mu_2}\boldsymbol{L}_{2}^{k}-\frac{\beta_2}{\mu_2}sgn(\boldsymbol{J}_2^{k})$.
		\STATE Update $\boldsymbol{Y}^{k+1}$,$\boldsymbol{Z}^{k+1}$ using Eq.~\eqref{eq:update_Y}.
		\STATE Update $\boldsymbol{L}_1^{k+1}$ and $\boldsymbol{L}_2^{k+1}$ as $\boldsymbol{L}_1^{k+1}=\boldsymbol{L}_1^{k}+\mu_1(\boldsymbol{J}_1^{k}-\boldsymbol{Y}^{k})$ and $\boldsymbol{L}_2^{k+1}=\boldsymbol{L}_2^{k}+\mu_1(\boldsymbol{J}_2^{k}-\boldsymbol{Y}^{k})$.
		\STATE $\mu_1\leftarrow\rho\mu_1$, $\mu_2\leftarrow\rho\mu_2$.
		\ENDFOR
		\STATE Post-compensation by $\boldsymbol{Y}\leftarrow\sqrt{tr(\boldsymbol{Y}^K)}\boldsymbol{Y}^K$.
		\ENSURE $\boldsymbol{Y}$.
	\end{algorithmic}  
\end{algorithm}

By setting $\eta_1=1/\mu_1$ for simplifying optimization, we obtain the final update process:
\begin{eqnarray}
\label{eq:update_J_1}
\boldsymbol{J}^{k+1}_1=\boldsymbol{Y}^{k}-\frac{1}{\mu_1}\boldsymbol{L}_1^{k}-\frac{\beta_1}{\mu_1}\boldsymbol{I}.
\end{eqnarray}


\textbf{Step 2: Update $\boldsymbol{J}_2$ while fixing other variables:}
Similar to Step 1, the irrelevant terms with $\boldsymbol{J}_2$ are ignored, and the objective function becomes: 

\begin{equation}
\small
\boldsymbol{J}_2^{k+1}=\arg\min_{\boldsymbol{J}_2^{k}} \beta_2||\boldsymbol{J}_2^{k}||_{1}+\frac{\mu_2}{2}||\boldsymbol{J}^{k}_2-\boldsymbol{Y}^{k}+\frac{1}{\mu_2}\boldsymbol{L}_2^{k}||^2_{F}.
\label{eq:obj_J2}
\end{equation}

Based on gradient descent optimization, $\boldsymbol{J}_2$ can be updated by:
\begin{equation}
\small
\label{eq:update_J2}
\boldsymbol{J}_2^{k+1}=\boldsymbol{J}^{k}_2-\eta_2[\beta_2\frac{\partial ||\boldsymbol{J}_2^{k}||_{1}}{\partial\boldsymbol{J}_2^{k}}+\mu_2(\boldsymbol{J}^k_2-\boldsymbol{Y}^k)+\boldsymbol{L}^k_2].
\end{equation}

In Eq.~\eqref{eq:update_J2}, we define $(\partial ||\boldsymbol{J}_2^{k}||_{1}/\partial \boldsymbol{J}_2^{k})= sgn(\boldsymbol{J}_2^{k})$, where $sgn(\cdot)$ is the sign function. With $\eta_2=1/\mu_2$, the $\boldsymbol{J}_2$ is updated by Eq.~\eqref{eq:update_J2_sign}.
\begin{eqnarray}
\label{eq:update_J2_sign}
\boldsymbol{J}^{k+1}_2=\boldsymbol{Y}^{k}-\frac{1}{\mu_2}\boldsymbol{L}_{2}^{k}-\frac{\beta_2}{\mu_2}sgn(\boldsymbol{J}_2^{k}).
\end{eqnarray}

The advantage of Eq.~\eqref{eq:update_J2_sign} is that no trainable parameters are required.
However, the negative updating direction of $sgn(\boldsymbol{J}_2^{k})$ indicates that all the elements in $\boldsymbol{Y}^k$ will be suppressed towards $0$ by subtracting a sign function, which is harmful to some critical elements for final recognition. An ideal solution is to only suppress the trivial elements in $\boldsymbol{Y}^k$ to achieve the sparsity constraint. 
Thus, we provide an alternative updating strategy, which is indicative of those important features in Sec.~\ref{sec:cam}.

\textbf{Step 3: Update $\boldsymbol{Y}$ while fixing other variables:}
By ignoring the irrelevant terms, $\boldsymbol{Y}$ is updated by Eq.~\eqref{eq:obj_J3}.
\begin{eqnarray}
\label{eq:obj_J3}
\footnotesize
\begin{split}
&\boldsymbol{Y}^{k+1}=\arg\min_{\boldsymbol{Y}^{k}} ||(\boldsymbol{Y}^{k})^{2}-\boldsymbol{A}||^2_{F}+\frac{\mu_1}{2}||\boldsymbol{J}^{k}_1-\boldsymbol{Y}^{k}+\frac{1}{\mu_1}\boldsymbol{L}_1^{k}||^2_{F}\\
&~~~~~~~~~+\frac{\mu_2}{2}||\boldsymbol{J}^{k}_2-\boldsymbol{Y}^{k}+\frac{1}{\mu_2}\boldsymbol{L}_2^{k}||^2_{F}.
\end{split}
\end{eqnarray}

By setting $\eta_3=1/(\mu_1+\mu_2)$, the gradient descent updating for $\boldsymbol{Y}$ is:
\begin{eqnarray}
\label{eq:update_J3_}
\begin{split}
\boldsymbol{Y}^{k+1}
=&\frac{\mu_1\boldsymbol{J}_1^k+\mu_2\boldsymbol{J}_2^k}{\mu_1+\mu_2}+\frac{\boldsymbol{L}_1^{k}+\boldsymbol{L}_2^{k}}{\mu_1+\mu_2}-\eta_3\boldsymbol{G},
\end{split}
\end{eqnarray}
where $\boldsymbol{G}$ is the updating direction in terms of square-root constraint:
\begin{eqnarray}
\label{eq:update_G}
\begin{split}
\boldsymbol{G}=\frac{\partial ||(\boldsymbol{Y}^{k})^{2}-\boldsymbol{A}||^2_{F}}{\partial \boldsymbol{Y}^k}.
\end{split}
\end{eqnarray}

We notice that directly employing $\boldsymbol{G}$ to update $\boldsymbol{Y}$ will lead to unstable convergence for iterative matrix square-root decomposition \cite{Higham2009}.
Therefore, we employ the efficient Newton Iteration, based on an auxiliary objective function $f_{N}(\boldsymbol{Y})=\boldsymbol{Y}^2-\boldsymbol{A}=0$, to calculate $\boldsymbol{G}$. Then, we have:
\begin{eqnarray}
\label{eq:NI_update}
\begin{split}
\boldsymbol{Y}^{k+1}&=\boldsymbol{Y}^k-\frac{f_{N}(\boldsymbol{Y}^k)}{f_{N}^{'}(\boldsymbol{Y}^k)}\\
&=\boldsymbol{Y}^k+\frac{1}{2}((\boldsymbol{Y}^k)^{-1}\boldsymbol{A}-\boldsymbol{Y}^k).
\end{split}
\end{eqnarray}

Since Eq.~\eqref{eq:NI_update} contains inefficient matrix inverse operation, a coupled updating form is further employed with matrix multiplication:

\begin{eqnarray}
\label{eq:NSI}
\begin{split}
\boldsymbol{Y}^{k+1}=&\boldsymbol{Y}^{k}+\frac{1}{2}\boldsymbol{Y}^{k}(\boldsymbol{I}-\boldsymbol{Z}^{k}\boldsymbol{Y}^{k}),\\
\boldsymbol{Z}^{k+1}=&\boldsymbol{Z}^{k}+\frac{1}{2}(\boldsymbol{I}-\boldsymbol{Z}^{k}\boldsymbol{Y}^{k})\boldsymbol{Z}^{k},
\end{split}
\end{eqnarray}
where $\boldsymbol{Z}^{k}$ is an auxiliary variable. Eq.~\eqref{eq:NSI} is also called as Newton-Schulz Iteration~\cite{Higham2009}.
If $||\boldsymbol{A}-\boldsymbol{I}||_{F}<1$, $\boldsymbol{Y}^{k}$ and $\boldsymbol{Z}^{k}$ can respectively converge to $\boldsymbol{A}^{1/2}$ and $\boldsymbol{A}^{-1/2}$, with quadratic speed.
Notably, $\boldsymbol{Z}^0=\boldsymbol{I}$.
Then, the $\boldsymbol{G}$ can be calculated efficiently by Eq.~\eqref{eq:NSI}, and the updating strategy towards square-root of Eq.~\eqref{eq:update_J3_} becomes:
\begin{eqnarray}
\label{eq:update_Y}
\begin{split}
\hat{\boldsymbol{Y}}^{k}~~~~=&\frac{\mu_1\boldsymbol{J}_1^k+\mu_2\boldsymbol{J}_2^k}{\mu_1+\mu_2}+\frac{\boldsymbol{L}_1^{k}+\boldsymbol{L}_2^{k}}{\mu_1+\mu_2},\\
\boldsymbol{Y}^{k+1}=&\hat{\boldsymbol{Y}}^{k}+\frac{1}{2}\hat{\boldsymbol{Y}}^{k}(\boldsymbol{I}-\boldsymbol{Z}^{k}\hat{\boldsymbol{Y}}^{k}),\\
\boldsymbol{Z}^{k+1}=&\boldsymbol{Z}^{k}+\frac{1}{2}(\boldsymbol{I}-\boldsymbol{Z}^{k}\hat{\boldsymbol{Y}}^{k})\boldsymbol{Z}^{k}.
\end{split}
\end{eqnarray}

Eq.~\eqref{eq:update_Y} first aggregates the low-rank $\boldsymbol{J}^k_1$ and sparse $\boldsymbol{J}^k_{2}$ as $\hat{\boldsymbol{Y}}^{k}$, and then update $\boldsymbol{Y}^{k+1}$ using efficient Newton Iteration.

\textbf{Step 4: Update multipliers:} The multipliers $\boldsymbol{L}_1$ and $\boldsymbol{L}_2$, as well as $\mu_1$ and $\mu_2$, are updated by:
\begin{eqnarray}
\label{eq:update_mul}
\begin{split}
\boldsymbol{L}_1^{k+1}=&\boldsymbol{L}_1^{k}+\mu_1(\boldsymbol{J}_1^{k}-\boldsymbol{Y}^{k}),\\
\boldsymbol{L}_2^{k+1}=&\boldsymbol{L}_2^{k}+\mu_2(\boldsymbol{J}_2^{k}-\boldsymbol{Y}^{k}),\\
\mu_1\leftarrow&\rho\mu_1;\mu_2\leftarrow\rho\mu_2,
\end{split}
\end{eqnarray}
where $\rho=1.1$ in this paper.

\subsection{Covariance Attention Mechanism}
\label{sec:cam}
\begin{figure}[t]
	\begin{center}
		\includegraphics[width=1\linewidth]{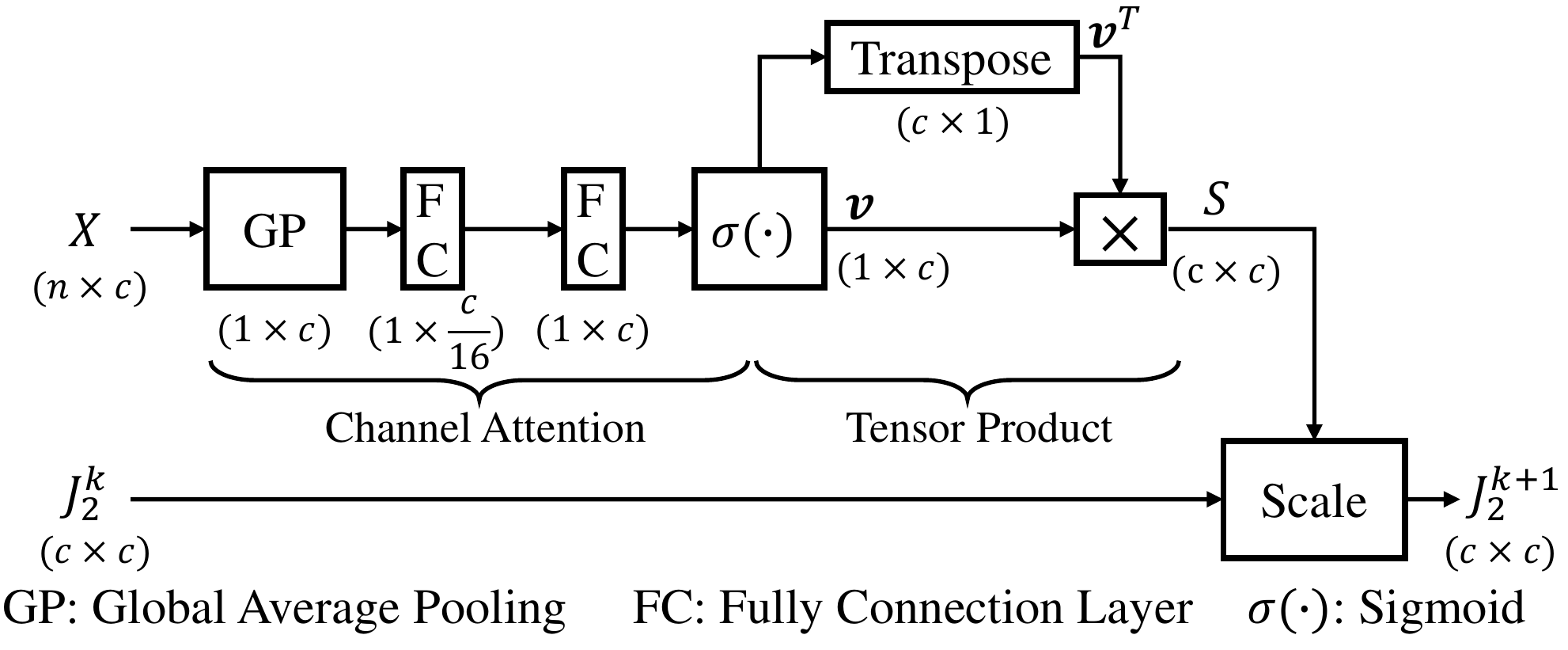}
	\end{center}
	\caption{A diagram of covariance attention mechanism.}
	\label{fig:bcam}
\end{figure}

In the updating step of $\boldsymbol{J}_2$ in Sec.~\ref{sec:opt}, we analyze that using Eq.~\eqref{eq:update_J2_sign} may damage some discriminative features.
Thus, a more comprehensive updating strategy is required, which is indicative of those discriminative features. 
To this end, we propose a covariance attention mechanism. 
As shown in Figure~\ref{fig:bcam}, the proposed covariance attention mechanism employs the channel attention vector $\boldsymbol{v}$ to generate a covariance attention map $\boldsymbol{S}=\boldsymbol{v}^{\top}\boldsymbol{v}$, whose size is the same as covariance matrix $\boldsymbol{J}_{2}^{k}$. 
Notably, $\boldsymbol{S}$ can be regarded as a dynamic weight to help $\boldsymbol{J}_{2}^{k+1}$ refine features from $\boldsymbol{J}_{2}^{k}$.
With the category label as supervision information \cite{Hu2018}, the loss gradients can optimize the inferred S so that the discriminative elements of $\boldsymbol{J}_{2}^{k}$ can be outputted.
In other words, the inferred $\boldsymbol{S}$ can reflect the importance of elements in $\boldsymbol{J}_{2}^{k}$.
Furthermore, to make $\boldsymbol{S}$ sparse, an inducing $l_1$ norm is used by minimizing $\mathcal{L}_{sr} = \frac{\beta_2}{c^2}\sum_{i}^{c}\sum_{j}^{c}|S_{i,j}|$, of which the sparsity is controlled by $\beta_2$.
As the attention value $S_{i,j}$ is non-negative, $\mathcal{L}_{sr}$ becomes:
\begin{eqnarray}
\label{eq:L_sp}
\mathcal{L}_{sr} = \frac{\beta_2}{c^2}\sum_{i=1}^{c}\sum_{j=1}^{c}S_{i,j}.
\end{eqnarray}

By using covariance attention mechanism and sparsity loss $\mathcal{L}_{sr}$, $\boldsymbol{S}$ has several advantages: a) sparsity; b) falling into $[0,1]$; c) indicating those important elements with large attention values.
Then, $(1-\boldsymbol{S})\boldsymbol{J}_{2}^{k}$ is exactly the redundant information in $\boldsymbol{J}_{2}^{k}$.
Removing $(1-\boldsymbol{S})\boldsymbol{J}_{2}^{k}$ has both effects of sparsity and redundancy reduction.
Finally, by replacing $sgn(\boldsymbol{J}_{2}^{k})$ with $(1-\boldsymbol{S})\boldsymbol{J}_{2}^{k}$, Eq.~\eqref{eq:update_J2_sign} becomes:
\begin{eqnarray}
\label{eq:update_J2_cam}
\boldsymbol{J}_{2}^{k+1}=\boldsymbol{Y}^{k}-\frac{1}{\mu_2}\boldsymbol{L}_{2}^{k}-\frac{\beta_2}{\mu_2}(1-\boldsymbol{S})\boldsymbol{J}_{2}^{k}.
\end{eqnarray}

\subsection{Implementation of MOMN layer}
Based on the above algorithm, the implementation of the proposed MOMN is shown in Figure~\ref{fig:pl}. 
In order to satisfy the condition of Newton-Schulz Iteration used for updating $\boldsymbol{Y}$ in Sec.~\ref{sec:opt}, the same pre-normalization and post-compensation in \cite{Li2018} are used by:
\begin{eqnarray}
\label{eq:pre_post_norm}
\boldsymbol{A}\leftarrow\frac{1}{tr(\boldsymbol{A})}\boldsymbol{A},~~~~~~~~\boldsymbol{Y}\leftarrow\sqrt{tr(\boldsymbol{Y})}\boldsymbol{Y}.
\end{eqnarray}
As $\boldsymbol{Y}$ is a symmetric matrix, we only output the upper triangular elements of $\boldsymbol{Y}$ to form a $c(c+1)/2$-dimensional vector, which is fed to the classifier.

Finally, the loss function for MOMN is:
\begin{eqnarray}
\label{eq:loss}
\mathcal{L} = \mathcal{L}_{cls}(\boldsymbol{Y},y) + \mathcal{L}_{sr}(\boldsymbol{S}),
\end{eqnarray}
where $\mathcal{L}_{cls}$ is the cross-entropy classification loss, and $y$ is the image label.
The classifier is omitted for simplifying.
$\mathcal{L}_{sr}$ is defined in Eq.~\eqref{eq:L_sp}, and $\boldsymbol{S}$ is the covariance attention map.
It should be noted that the implementation of MOMN layer contains only matrix multiplication, which is friendly to GPU acceleration.

\section{Experiments}
\label{sec:exp}
In this section, we first provide the detailed experimental setting and then conduct several ablation studies to evaluate each component in MOMN.
Finally, MOMN is compared with the state-of-the-art methods on five benchmarks with different backbones.

\subsection{Experimental Setting}
\noindent\textbf{Datasets} We evaluate the proposed MOMN on five benchmarks:
Caltech-UCSD birds (CUB) \cite{wah2011caltech}, Standford Cars (Cars) \cite{Krause2013}, FGVC-aircraft (Aircraft) \cite{Maji2013}, Standford Dogs (Dogs) \cite{khosla2011novel}, and MPII Human Pose Dataset (MPII) \cite{Pishchulin2014}. 
The first four datasets are used for image classification, which are commonly used in FGVC.
CUB contains 11,788 images of 200 species of birds, which have subtle inter-class differences.
The whole images are divided into two parts: 5,994 images for training, and 5,794 images for testing.
Similarly, Cars contains 8,144 training images and 8,041 testing images from 196 classes of cars.
Aircraft contains 6,667 images for training and 3,333 images for testing from 100 classes, and Dogs contains 12,000 images for training and 8,580 images for testing from 120 classes.
The last dataset is about fine-grained action recognition, which contains 15,205 images from 393 action classes, to evaluate the model generalization on different tasks.
The dataset is split into train, val and test sets by authors \cite{Gkioxari2015}, with respectively 8,218, 6,987 and 5,708 samples.
Since MPII only provides the pose estimation evaluation on the test set, the comparisons are commonly conducted on the val set \cite{girdhar2017attentional}.
The detail information of the above benchmarks is given in Table~\ref{tab:datasets}.

\begin{table}
\begin{center}
\caption{Details of experimental datasets. The train/test indicates the image number of respective data split.} \label{tab:datasets}
\begin{tabular}{lccc}
  \hline
  Datasets &Class Number&Train&Test \\
  \hline
  \hline
  Birds~\cite{wah2011caltech}&200&5,994&5,794\\
  Cars~\cite{Krause2013}&196&8,144&8,041\\
  Aircraft~\cite{Maji2013}&100&6,667&3,333\\
  Dogs~\cite{khosla2011novel}&120&12,000&8,580\\
  MPII~\cite{Pishchulin2014}&393&8,218&6,987\\
  \hline
\end{tabular}
\end{center}
\end{table}

\noindent\textbf{Implementation Details}
Following the evaluation setting in~\cite{Simon2017}, the input images are cropped to $448\times 448$, and the backbone networks are pre-trained on ImageNet. The horizontal flipping is used for data augmentation during training.
Taking ResNet-101~\cite{he2016deep} as backbone, the reduction layer in Figure~\ref{fig:pl} can reduce the channels of the last layer features from $2,048$ to $256$, thereby producing $256\times 256$ covariance matrix $\boldsymbol{A}$.
A two-step training strategy~\cite{Simon2017} is applied to train the model.
First, the pre-trained backbone network is fixed, and the rest part is trained with a large learning rate (\emph{lr}=0.01). Then, the whole network is fine-tuned with a small learning rate (\emph{lr}=0.001).
During training, a mini-batch contains $16$ samples, and SGD with the momentum of 0.9 for 180 epochs is used as the optimizer.
The hyper-parameter settings are $\beta_1=0.5$, $\beta_2=1$, and $K=5$, which will be analysed in ablation studies.
The Lagrange parameters are fixed with $\mu_1=\mu_2=10$ for all cases.
For testing, the averaged score of an image and its horizontal flip is reported as the final prediction.

\begin{table}
\begin{center}
\caption{Evaluation of different architectures for dimension reduction layer. A block indicates the combination of conv+bn+relu. The top-1 accuracy is reported.} \label{tab:rl}
\begin{threeparttable}
\begin{tabular}{lcccc}
  \hline
  Architecture&Conv&1$\times$Block$\dag$&2$\times$Block&3$\times$Block\\
  \hline
  \hline
  Resnet-50& 87.91 & 88.32 &88.30&88.25\\
  \hline
  Resnet-101& 89.22 & 89.41 &89.25&89.31\\
  \hline
\end{tabular}
\begin{tablenotes}
	\footnotesize
	\item $\dag$: Block: Conv+BN+ReLU; 
\end{tablenotes}
\end{threeparttable}
\end{center}
\end{table}

\subsection{Ablation Studies}
In this section, we give several detailed analyses about the MOMN. The experiments are all performed on GTX1080ti with the backbone network of ResNet-50 and ResNet-101.
Except for the classifier, the rest experiment settings are consistent with iSQRT \cite{Li2018}, ~\emph{i.e.,} MOMN uses the softmax classifier rather than SVM used in iSQRT.

\noindent\textbf{Evaluation of reduction layer}
To evaluate the effect of dimension reduction layer, we conduct several evaluations for different settings of the dimension reduction layer, and summarize the related results in Table~\ref{tab:rl}.
For the proposed MOMN, we use a simple conv+bn+relu as the dimension reduction layer before the bilinear pooling, which is widely used in existing works \cite{li2017second,Li2018}.
As shown in Table~\ref{tab:rl}, using the Blocks that consist of conv.+bn+relu all achieves higher performance than using a simple Conv layer. Further, using a single Block obtains the best performance.
As a consequence, it is enough to use a block of conv.+bn+relu to reduce the dimension of backbone features.

\begin{figure}
	\begin{center}
		\includegraphics[width=1\linewidth]{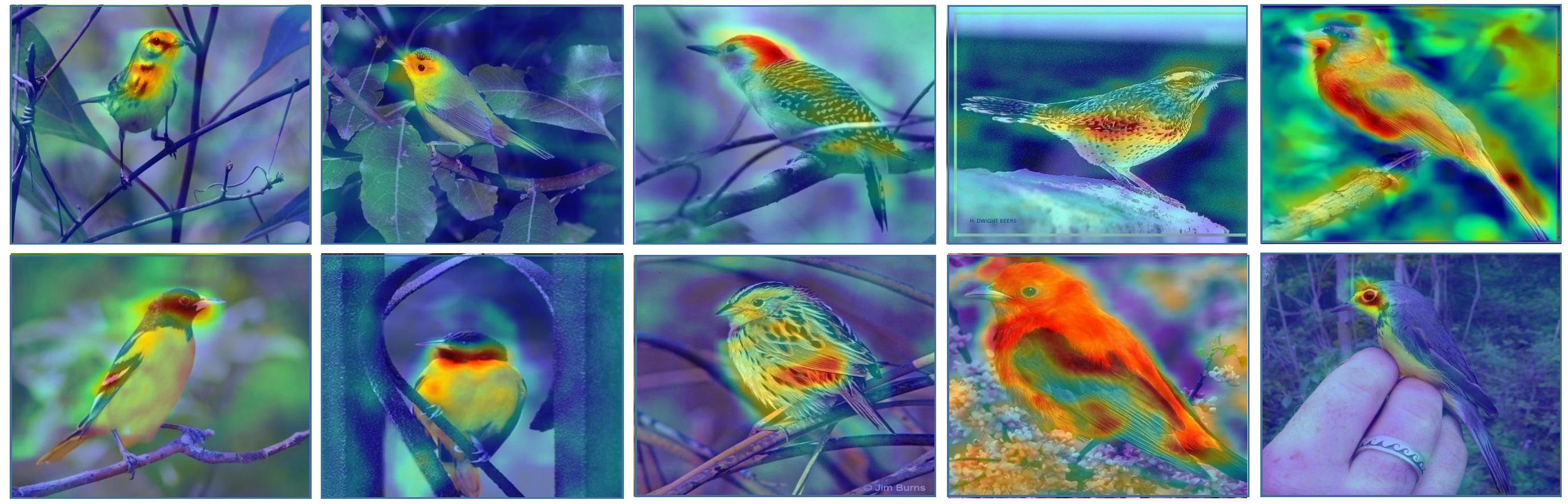}
	\end{center}
	\caption{Some visualized feature maps of maximum response in the generated attention vector $\boldsymbol{v}$. Usually, the birds head and abdomen are considers as informative regions.}
	\label{fig:att}
\end{figure}

\begin{figure}
	\begin{center}
		\includegraphics[width=1\linewidth]{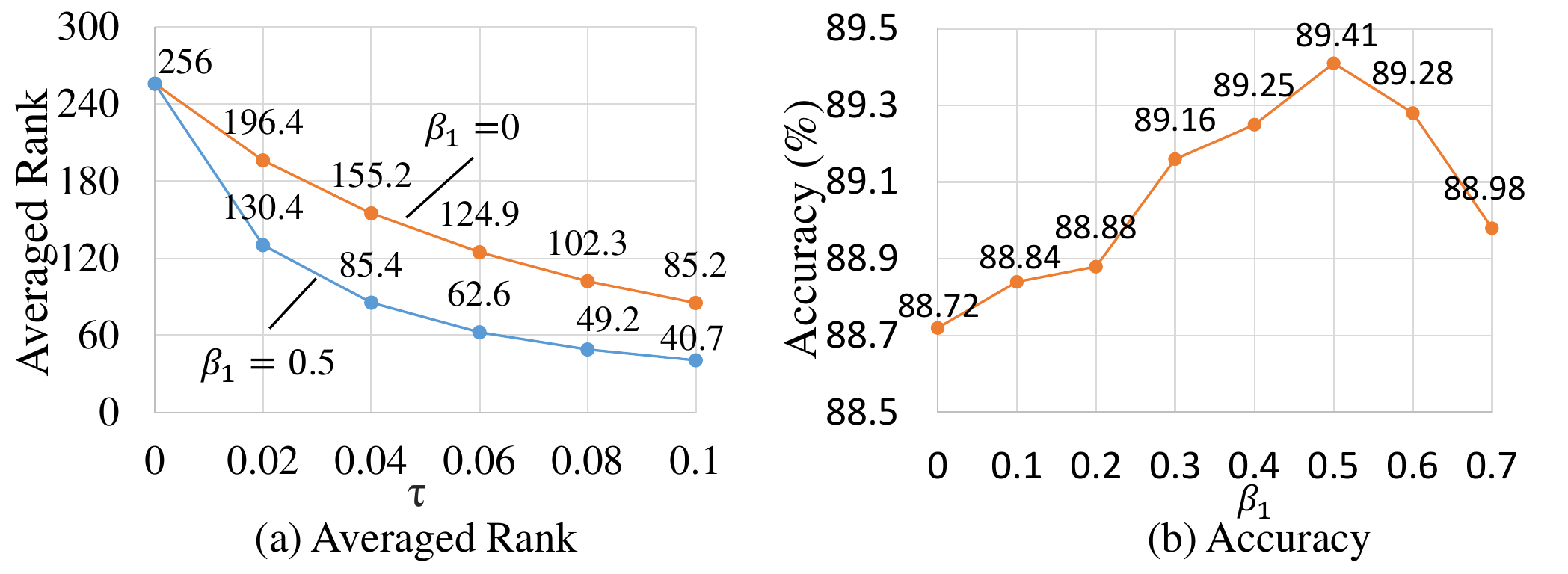}
	\end{center}
	\caption{Evaluation of low-rank effect by MOMN on CUB, in terms of $\tau$ and $\beta_1$.}
	\label{fig:exp_r}
\end{figure}

\noindent\textbf{Effect of low-rank constraint}
To demonstrate the effectiveness of low-rank constraint, we compare the averaged rank of normalized $\boldsymbol{Y}$ between the settings with $\beta_1=0$ and $\beta_1=0.5$. The related results are shown in Figure~\ref{fig:exp_r}.
As the nuclear norm $||\boldsymbol{Y}||_{*}$ is an approximated rank norm by suppressing singular values of matrix components, we define the approximated rank $r(\boldsymbol{Y})\approx num(\{\lambda_i|\lambda_i>\tau\})$ by counting the number of singular values that are higher than $\tau$.
$\tau$ is a small value and used to filter out the suppressed matrix components.
As shown in Figure~\ref{fig:exp_r} (a), some matrix components have been exactly suppressed after using the MOMN, leading to a low-rank $\boldsymbol{Y}$. 
For example, when $\tau\geq 0.04$, nearly half of $r(\boldsymbol{Y})$ is suppressed by MOMN.
The above analysis demonstrates that our MOMN can effectively reduce the rank of regularized bilinear features. 

$\beta_1$ is a critical parameter to balance the effect of low-rank constraint in the objective function of Eq.~\eqref{eq:obj_f}.
A larger $\beta_1$ corresponds to a stronger low-rank constraint.
As displayed in Figure~\ref{fig:exp_r} (b), the performance increases at the beginning.
Thus, it is effective by removing the redundant information in $\boldsymbol{Y}$ to obtain discriminative image representation.
Then, the performance drops seriously, when $\beta_1$ is larger than $0.5$.
This tells that a too large $\beta_1$ is harmful to the principle components in $\boldsymbol{Y}$, due to an over-strong low-rank constraint.
Therefore, we set $\beta_1$=$0.5$ in the following experiments.

\begin{table}
\begin{center}
\caption{Effect of each regularizer. ACC. is the classification accuracy, and Time shows the inference time. MOMN\_sgn and MOMN indicate that using the plain sign function and covariance attention for sparsity updating, respectively. $\dagger$ indicates using sign function.} \label{tab:eval_norm}
\resizebox{1\columnwidth}{!}{
\begin{tabular}{c|ccc|c|c}
  \hline
  \multirow{2}{*}{Config.}& \multicolumn{3}{c|}{Matrix Normalizations}& \multirow{2}{*}{Time (\emph{ms})} &\multirow{2}{*}{ACC. (\%)}\\
  \cline{2-4}
  & square-root & low-rank &  sparsity&\\
  \hline
  \hline
  Baseline&&&&-&84.11\\
  \hline
  \multirow{4}{*}{SON}&$\surd$&&&0.54&$\textbf{88.12}$\\
  &&$\surd$&&$\textbf{0.25}$&86.19\\
  &&&$\surd$&0.38&85.41\\
  \hline
  \multirow{3}{*}{TON}&$\surd$&$\surd$&&0.82&$\textbf{88.91}$\\
  &$\surd$&&$\surd$&0.94&88.58\\
  &&$\surd$&$\surd$&$\textbf{0.61}$&87.63\\
  \hline
  MOMN\_sgn&$\surd$&$\surd$&$\dagger$&$\textbf{1.30}$&89.20\\
  MOMN&$\surd$&$\surd$&$\surd$&1.31&$\textbf{89.41}$\\
  \hline
  
\end{tabular}}
\end{center}
\end{table}

\begin{figure}
	\begin{center}
		\includegraphics[width=1\linewidth]{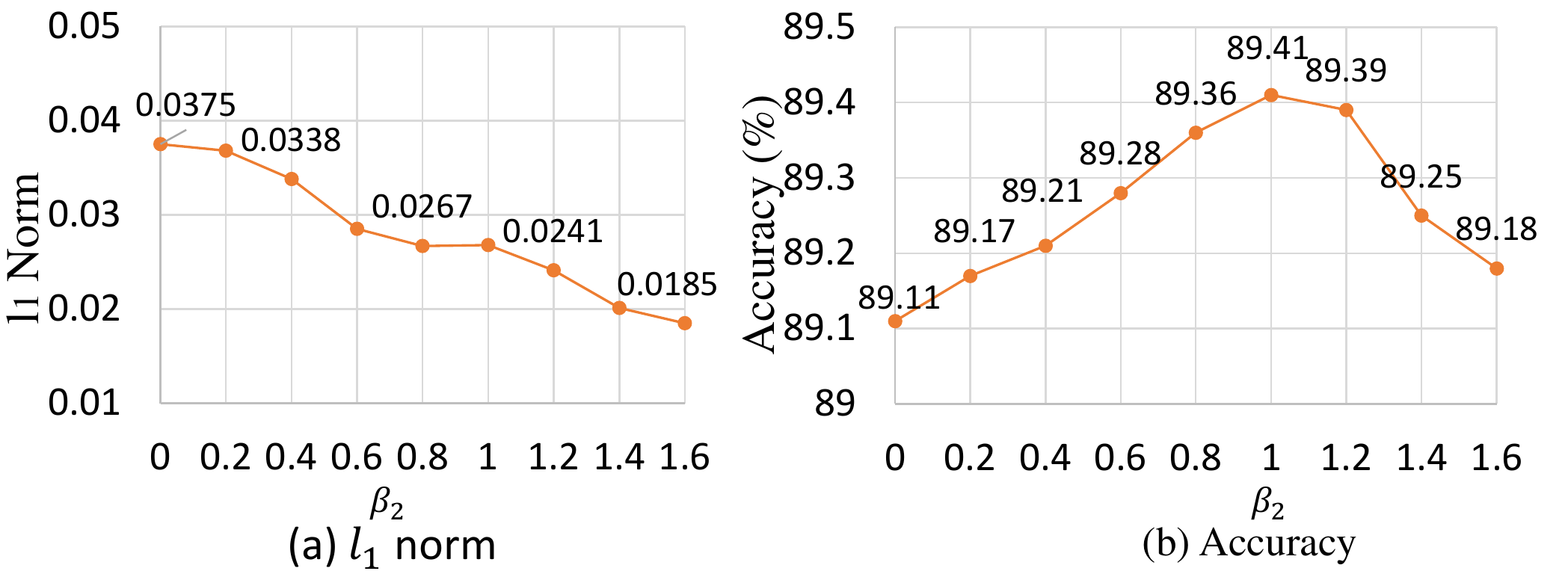}
	\end{center}
	\caption{Evaluation of sparsity constraint of MOMN on CUB. Both the $l_1$ norm and accuracy are reported with varying $\beta_2$.}
	\label{fig:exp_s}
\end{figure}

\noindent\textbf{Effect of sparse constraint}
The effect of sparsity constraint is also explored by evaluating different $\beta_2$.
From Figure~\ref{fig:exp_s} (a), the sparsity term $||\boldsymbol{Y}||_{1}$ is effectively reduced, which shows that the normalized $\boldsymbol{A}$ becomes sparse with MOMN.
From Figure~\ref{fig:exp_s} (b), when $\beta_2$ increases from $0$ to $1.0$, the accuracy is improved from $89.11\%$ to $89.41\%$. 
This proves the positive effect of sparsity in bilinear representation learning.
However, the performance will drop seriously when further increasing $\beta_2$. The reason is that a larger $\beta_2$ leads to a stronger sparsity constraint on the bilinear features. In the rest experiments, $\beta_2$ is fixed to $1.0$.

\noindent\textbf{Comparison of three matrix regularizers} 
In this part, we evaluate the effect of each regularizer in MOMN and summarize the related results in Table~\ref{tab:eval_norm}. Note that the baseline performance is generated by original bilinear pooling with $l_2$ normalization. 
First, we evaluate the effect of each constraint with Single-Objective Normalization (SON). From Table~\ref{tab:eval_norm}, we can observe that the square-root constraint achieves the highest accuracy among all three constraints, \emph{e.g.,} about 4\% higher than the baseline accuracy. The larger improvement testifies the importance and necessity of the square-root constraint for bilinear representation learning. Besides, the low-rank and sparsity constraints achieve about 2\% and 1.3\% improvement on the baseline performance.
Then, we continue to evaluate different combinations of constraints, denoted as Two-Objective Normalization (TON). As shown in Table~\ref{tab:eval_norm}, two-objective settings achieve higher performance than all single-objective settings. 
Especially, the TON with square-root and low-rank constraints gets the highest accuracy, which obtains about 4.8\% improvement over the baseline. 
Finally, we show the performance by using all three matrix constraints. From Table~\ref{tab:eval_norm}, MOMN outperforms baseline and SON by about 5.3\% and 1.3\% improvements.
It indicates the importance of multi-objective normalization in bilinear features. Therefore, we can conclude that the three matrix constraints and multi-objective normalization are both useful for bilinear representation learning.

\begin{table}
\begin{center}
\caption{Effects of sequentially adding each regularizer, where SR, LR, and SP correspond to Square-Root, Low-Rank, and Sparsity, respectively. SR$\rightarrow$LR$\rightarrow$SP indicates that sequentially applying the three regularizers to a bilinear feature.} \label{tab:joint_opt}
\resizebox{1\columnwidth}{!}{
\begin{tabular}{ccccc}
  \hline
  Regularizer&SR&SR$\rightarrow$LR&SR$\rightarrow$LR$\rightarrow$SP&MOMN\\
  \hline
  \hline
  ACC. (\%)&88.1 & 87.5 &86.8&89.41\\
  \hline
\end{tabular}}
\end{center}
\end{table}

\noindent\textbf{Covariance attention vs. sign function}
For the sparsity updating step in Sec.~\ref{sec:opt}, we provide two updating strategies: covariance attention mechanism and sign function. Therefore, we present a comparison between these two updating strategies. From Table~\ref{tab:eval_norm}, using covariance attention obtains $0.2\%$ improvement over sign function, with ignorable time sacrifice of 0.01 \emph{ms}.
Further, we provide some visualized results to illustrate the effectiveness of covariance attention. 
Since the covariance attention is based on the channel attention mechanism by $\boldsymbol{S}=\boldsymbol{v}^T\boldsymbol{v}$, as shown in Figure~\ref{fig:bcam}, we show those feature maps in $\boldsymbol{X}$ that are indicated by the maximum values in the attention vector $\boldsymbol{v}$.
The results are given in Figure~\ref{fig:att}.
From the results, we can see that the covariance attention mechanism focuses on those object regions,~\emph{i.e.,} the birds head and abdomen, where the discriminative information is contained.
Thus, using the covariance attention mechanism can make the generated representation compact and discriminative.

\begin{table}
\begin{center}
\caption{Comparison of forward and backward time (\emph{s}) and classification accuracy (\%) used of different matrix normalization layers. The accuracy of iSQRT in parentheses is the averaged accuracy with Softmax classifier.} \label{tab:eval_time}
\resizebox{1\columnwidth}{!}{
\begin{tabular}{l|c|c|c|c|c}
  \hline
  Method& Bottleneck & Constraints & FT & BT& ACC.\\
  \hline
  \hline
  iBCNN\cite{lin2017improved}&SVD&Single&12.38&1.20&85.8\\
  \hline
  G$^2$DeNet\cite{Wang2017}&SVD&Single&8.11&0.72&87.1\\
  \hline
  MPN-COV\cite{li2017second}&EIG&Single&4.64&0.88&-\\
  \hline
  iSQRT\cite{Li2018}&NS Iter&Single&0.61&0.32&88.7(88.2)\\
  \hline
  \multirow{3}{*}{MOMN}&\multirow{3}{*}{ADMM}&Single&$\textbf{0.54}$&$\textbf{0.32}$&88.1\\
  &&Two&0.82&0.33&88.9\\
  &&Three&1.31&0.34&$\textbf{89.4}$\\
  \hline
\end{tabular}}
\end{center}
\end{table}

\begin{figure}
	\begin{center}
		\includegraphics[width=1\linewidth]{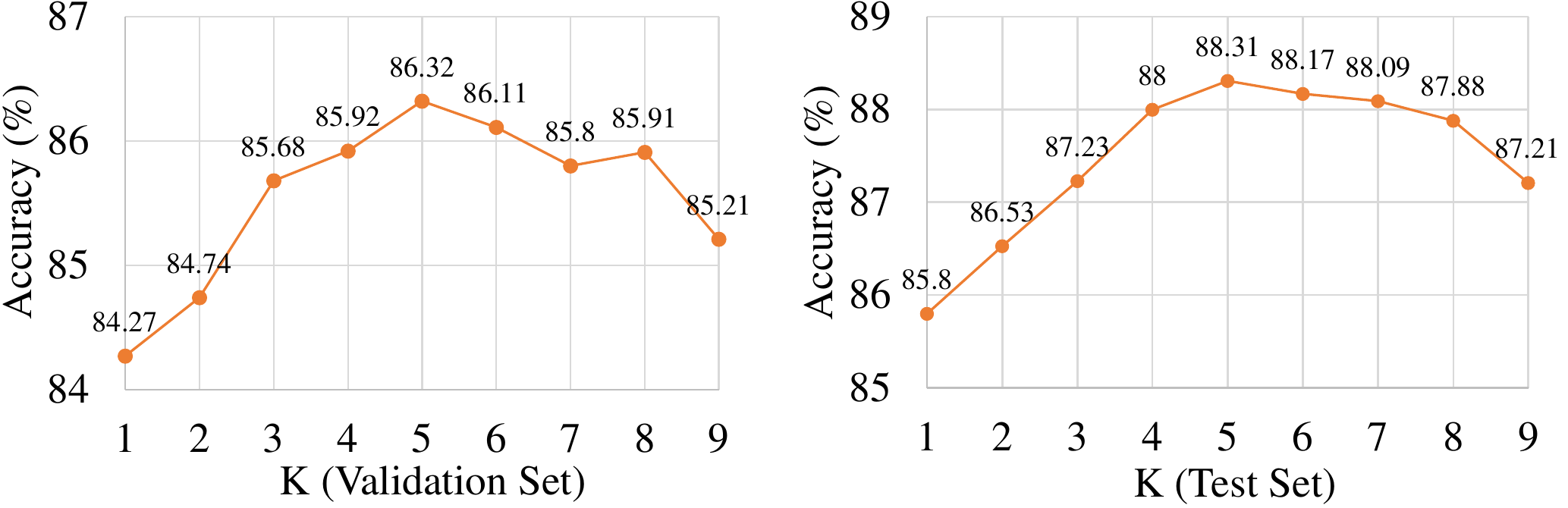}
	\end{center}
	\caption{Impact of $K$ of MOMN on CUB dataset. Both validation and Test sets are evaluated. The backbone is the Resnet-50.}
	\label{fig:exp_k}
\end{figure}

\noindent\textbf{Necessity of joint multi-objective normalization}
Since the three regularizers are important for bilinear representation learning, we then show that the joint optimization is necessary for multi-objective normalization. 
In Table~\ref{tab:joint_opt}, we first minimize the square-root regularizer (SR) of $||\boldsymbol{Y}^2-\boldsymbol{A}||_2^2$ to make $\boldsymbol{Y}\rightarrow\boldsymbol{A}^{1/2}$, and then optimize $||\boldsymbol{Y}||_{*}$ (LR) to make $\boldsymbol{Y}$ low rank. Finally, the sparse regularizer $||\boldsymbol{Y}||_1$ (SP) is used to make $\boldsymbol{Y}$ sparse.
Different from the proposed joint multi-objective normalization in MOMN, such a direct adding strategy is simple, without considering the impact between different constraints,~\emph{e.g.,} no alternate updating and no Lagrange multipliers as penalties.
Specifically, optimizing the latter regularizer will corrupt the feature structure normalized by the previous one, \emph{e.g.,} adding low-rank to square-root reduces its performance from 88.1\% to 87.5\%. 
Therefore, it is crucial to employ a unified framework to optimize all three regularizers jointly.

\noindent\textbf{Effect of Iteration Number}
To determine the optimal $K$ in the MOMN, we further split the training set of CUB into train/val parts, according to $3:2$ for each category.
Then, we evaluate the effects of different $K$ on the validation set.
Finally, different $K$ is also evaluated on the official train/test set to prove the generalization.
The results are reported in Figure~\ref{fig:exp_k}. 
It can be found that a larger $K$ corresponds to higher performance.
However, when $K>5$, the accuracy drops.
The reason may be that too many iterations for gradient descent updating with fixed steps may miss the optimal global solution.
Hence, we set $K=5$ in this work.

\begin{figure}
	\begin{center}
		\includegraphics[width=1\linewidth]{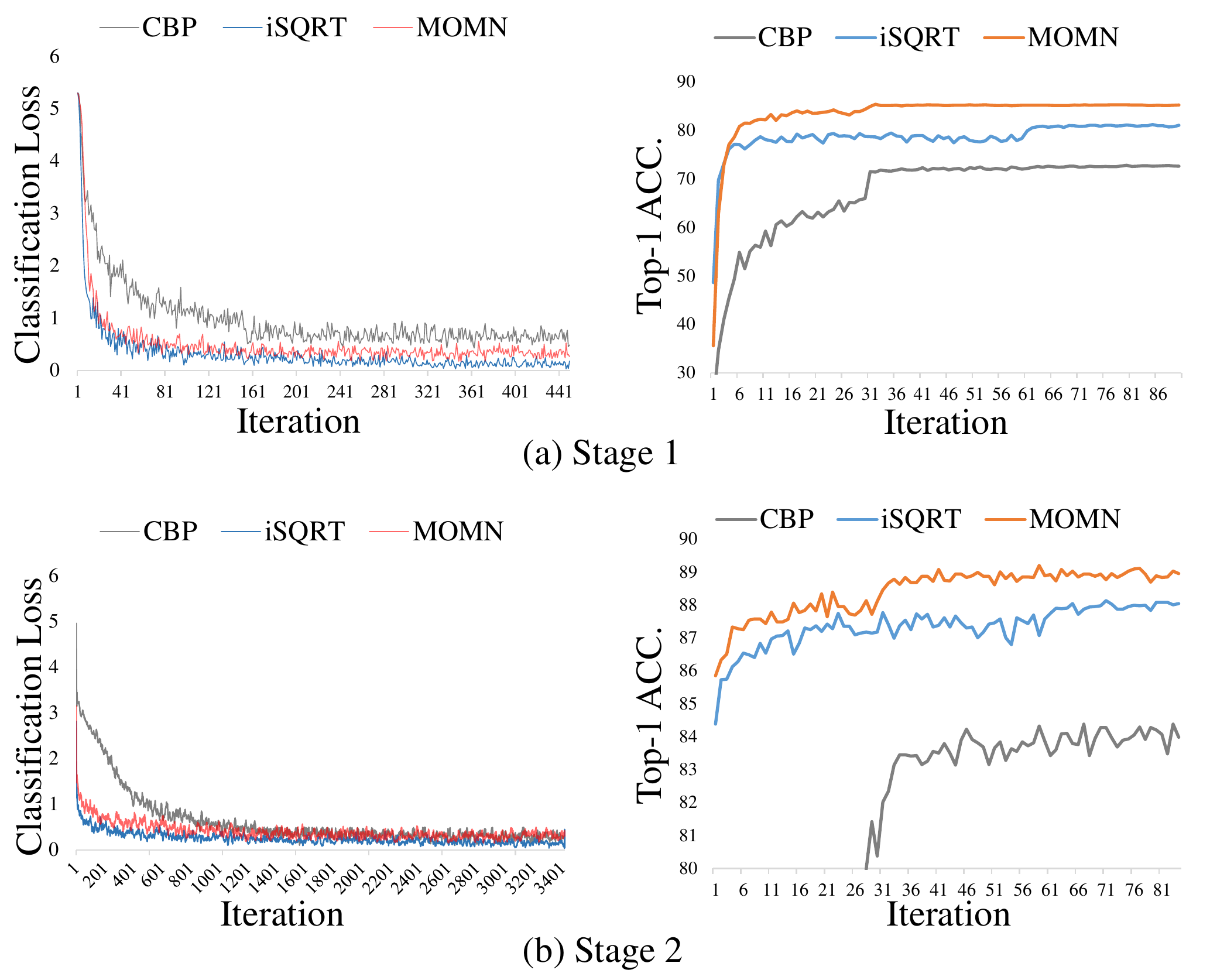}
	\end{center}
	\caption{The comparison of Loss-Iteration and ACC-Iteration curves of different normalization methods on CUB dataset. The classification loss is the cross-entropy loss.}
	\label{fig:training}
\end{figure}

\noindent\textbf{Analysis of training convergence}
Figure~\ref{fig:training} shows the loss and accuracy curve during two-stage training.
In stage one, the backbone network is fixed, and in stage two, the whole network is fine-tuned.
We compare MOMN with two representative methods of CBP \cite{gao2016compact} and iSQRT \cite{Li2018}.
From Figure~\ref{fig:training}, several conclusions can be obtained.
First, both iSQRT and MOMN have faster and better convergence than CPB during both training stages.
The better performance proves that the matrix square-root normalization has a stronger stabilization effect than $l_2$ normalization of CBP.
Second, MOMN obtains higher accuracy than iSQRT with approximated training error, because MOMN has a better generalization than iSQRT.
The reason is that the extra low-rank and sparsity regularizers can effectively remove the trivial components in the bilinear features, thereby making them discriminative and generalized.
Third, the accuracy convergence of MOMN in the second stage is faster than iSQRT.
This further proves the training stability of multi-objective normalization in FGVC.

\begin{figure*}
	\begin{center}
		\includegraphics[width=0.95\linewidth]{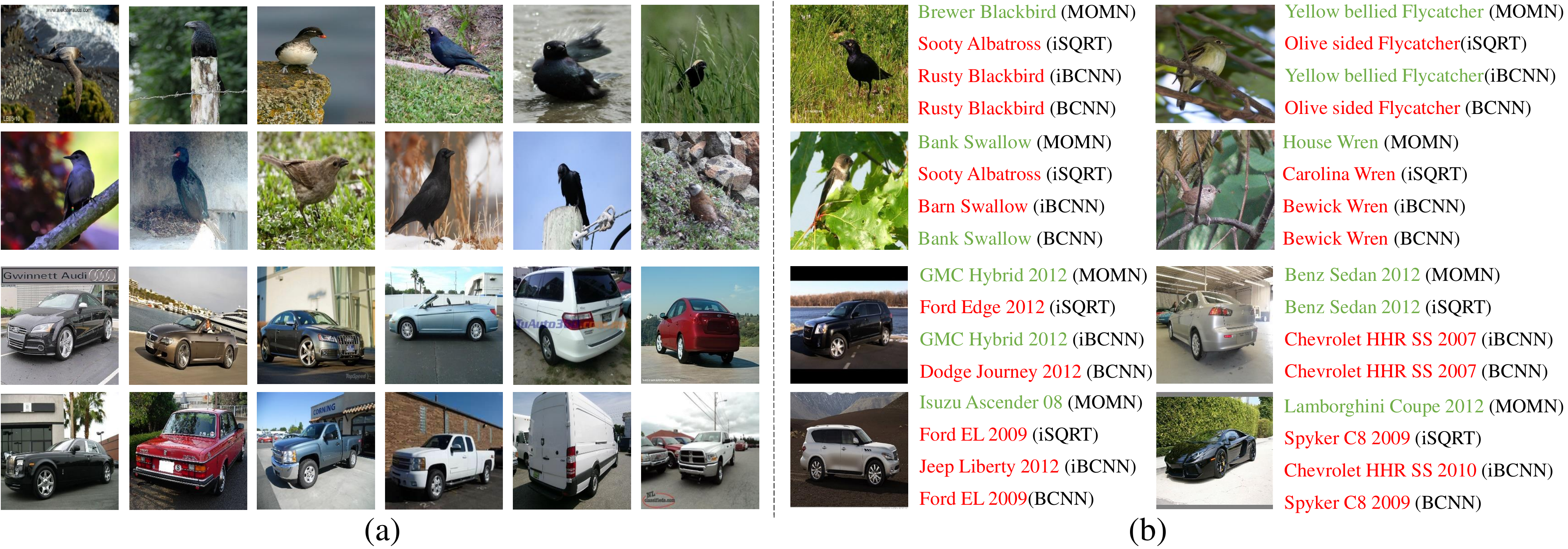}
	\end{center}
	\caption{The part (a) shows some examples that are correctly recognized by the proposed MOMN, but cannot be handled by the related works of BCNN \cite{Lin2015}, iBCNN \cite{lin2017improved}, and iSQRT \cite{ Li2018}. The part (b) gives some recognition results of the four methods. From the results, those images with confusing background appearance and inapparent characteristics are usually hard for previous methods. This indicates that the low-rank and sparsity constraints enable MOMN to remove those trivial components and avoid the over-fitting problem.}
	\label{fig:vis}
\end{figure*}

\noindent\textbf{Efficiency analysis of MOMN layer}
We target to propose an efficient multi-objective normalization for bilinear representation.
Thus, the time cost of each component during the forward propagation of MOMN is given.
From Table~\ref{tab:eval_norm}, the matrix square-root decomposition spends the most time in MOMN, followed by sparsity and low-rank constraints.
Since MOMN is implemented with only matrix multiplication, the time costs of all three regularizers are small.
Notably, the extra time-consuming by updating the Lagrange multiplier is ignorable, which is about $0.14$ \emph{ms}.

Furthermore, we compare the proposed MOMN with existing normalization methods,~\emph{i.e.,} G$^2$DeNet \cite{Wang2017}, Impro. B-CNN (SVD version) \cite{lin2017improved}, MPN-COV \cite{li2017second}, and iSQRT \cite{Li2018}.
Both forward propagation (FP) and back propagation (BP) are analyzed, of which the results are summarized in Table~\ref{tab:eval_time}.
Compared with those SVD or EIG based-methods, MOMN and iSQRT give superior implementation efficiency in terms of both FP and BP. 
Compared with iSQRT, our MOMN obtains comparable efficiency under the single-objective setting.
Notably, the reason for higher accuracy of iSQRT under the single constraint is that it utilizes the SVM as the classifier, which is much stronger than the simple Softmax classifier in MOMN.
Once replacing the SVM classifier with the softmax classifier, the accuracy for iSORT drops from 88.7\% to 88.2\%, which is comparable with the proposed MOMN.
After using multi-objective setting, MOMN obtains higher performance with little time addition.
In summary, the proposed MOMN achieves a good balance between efficient implementation and multi-objective normalization.

\begin{table}
\begin{center}
\caption{Evaluation of MOMN on CUB, Cars, and Aircraft datasets. The evaluation metric is classification accuracy ($\%$). VGG-D contains 16 and 19 layer settings. Densenet-D contains 161 and 201 layer settings. ReNet-D contains 50 and 101 layer settings. * indicates that the methods are based on part localization.} \label{tab:fgvc_obj}
\resizebox{1\columnwidth}{!}{
\begin{tabular}{lcccc}
  \hline
  Methods&Backbone& CUB & Cars& Aircraft \\
  \hline
  \hline
  CBP\cite{gao2016compact}&\multirow{14}{*}{VGG-D}&84.3&91.2&84.1\\
  LR-BCNN\cite{kong2017low}&&84.2&90.9&87.3\\
  $\alpha$-pooling\cite{Simon2017}&&85.3&-&85.5\\
  KP\cite{cui2017kernel}&&86.2&92.4&86.9\\
  G$^2$DeNet\cite{Wang2017}&&87.1&92.5&89.0\\
  Impro. BCNN\cite{lin2017improved}&&85.8&92.0&88.5\\
  HIHCA\cite{cai2017higher}&&85.3&91.7&88.3\\
  MoNet\cite{gou2018monet}&&86.4&91.8&89.3\\
  FBP \cite{yu2018hierarchical}&&85.7&92.1&88.7\\
  Grass. Pool\cite{wei2018grassmann}&&85.8&$\boldsymbol{92.8}$&89.8\\
  iSQRT\cite{Li2018}&&87.2&92.5&90.0\\
  \textbf{MOMN}&&$\boldsymbol{87.3}$&$\boldsymbol{92.8}$&$\boldsymbol{90.4}$\\
  \hline
  CBP\cite{gao2016compact}&\multirow{4}{*}{Densenet-D}&83.1&87.9&83.2\\
  PC\cite{dubey2018pairwise}&&86.9&92.9&89.2\\
  iSQRT\cite{Li2018}&&88.5&92.1&90.9\\
  \textbf{MOMN}&&$\boldsymbol{89.7}$&$\boldsymbol{93.7}$&$\boldsymbol{91.8}$\\
  \hline
  *S3N\cite{ding2019selective}&\multirow{10}{*}{Resnet-D}&88.5&94.7&92.8\\
  *MGE-CNN\cite{zhang2019learning}&&89.4&39.9&-\\
  CBP\cite{gao2016compact}&&81.6&88.6&81.6\\
  KP\cite{cui2017kernel}&&84.7&91.1&85.7\\
  PC\cite{dubey2018pairwise}&&85.6&92.5&85.8\\
  DCL\cite{Chen_2019_CVPR}&&86.2&93.4&89.9\\
  iSQRT\cite{Li2018}&&88.7&93.3&91.4\\
  \textbf{MOMN} (D=50)&&88.3&93.2&90.3\\
  \textbf{MOMN} (D=101)&&89.4&93.9&92.0\\
  \textbf{MOMN} (D=101\_32x8d)&&$\boldsymbol{89.8}$&$\boldsymbol{94.2}$&$\boldsymbol{92.2}$\\
  \hline
\end{tabular}}
\end{center}
\end{table}

\begin{figure*}
	\begin{center}
		\includegraphics[width=0.95\linewidth]{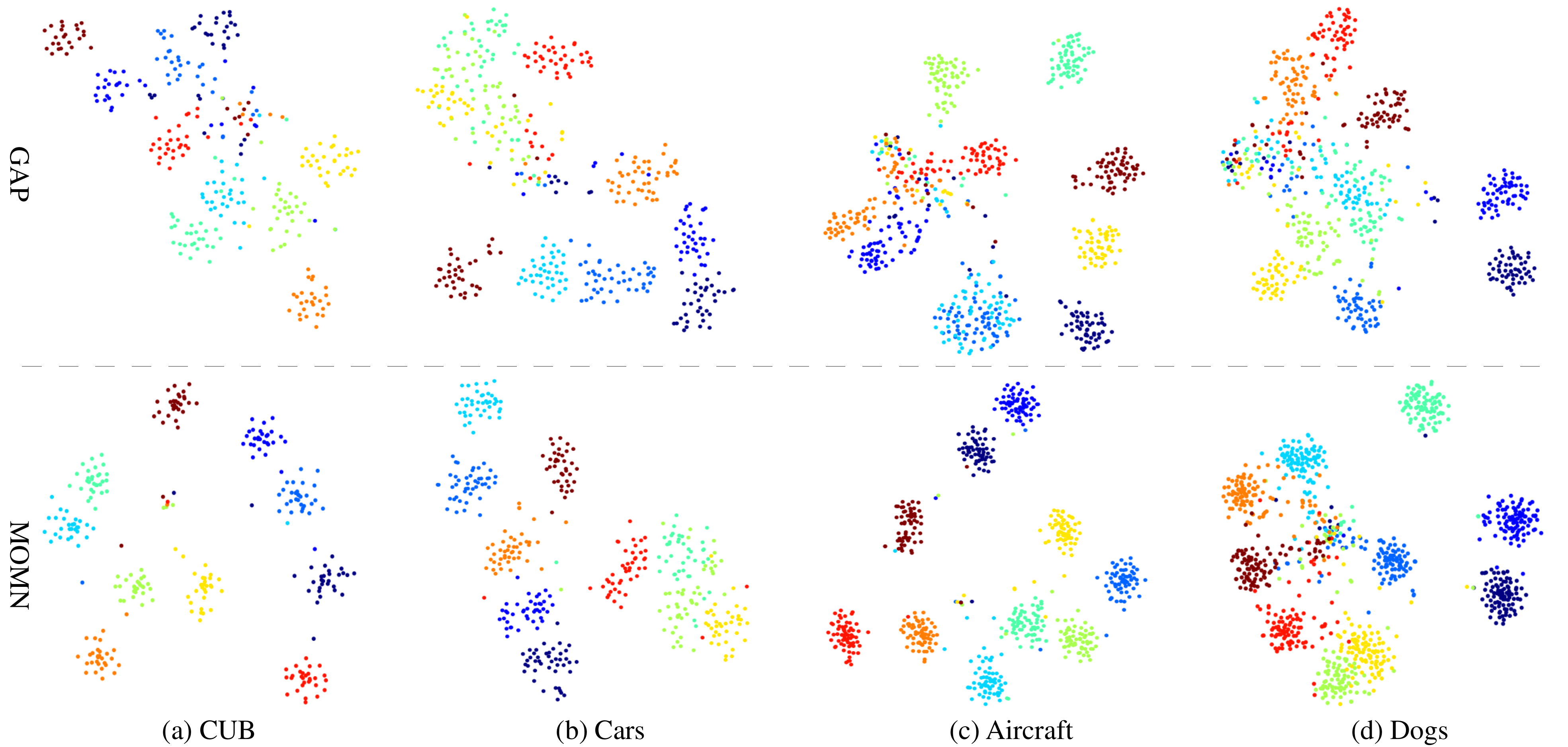}
	\end{center}
	\caption{The feature distributions of the global average pooling and MOMN in four datasets, which are obtained by t-SNE method. In each dataset, random $10$ classes are selected for visualization.}
	\label{fig:tsne}
\end{figure*}

\noindent\textbf{Visualization results}
Finally, we visualize the feature distributions obtained from the global average pooling (GAP) and MOMN to show the superiority of normalized bilinear features.
From Figure~\ref{fig:tsne}, it can be seen that the feature distributions of MOMN have larger inter-class differences than GAP, as well as more compact clusters.
Notably, all the evaluated datasets contain fine-grained categories, which are visually similar to each other.
Thus, the simple GAP will miss some imperceptible visual clues via spatial summation, leading to blurred decision boundary.
With effective normalizations, MOMN can model the second-order information to capture the vital clues about the inter-class difference. 
Specifically, the low-rank and sparsity regularizers in MOMN play important roles in generating discriminative features in Figure~\ref{fig:tsne}, via redundancy reduction and alleviating over-fitting.

Further, we also provide some image examples in Figure~\ref{fig:vis}.
These images are correctly recognized by MOMN, but the previous BCNN \cite{Lin2015}, iBCNN \cite{lin2017improved}, and iSQRT \cite{ Li2018} cannot handle.
Notably, these images are usually with confusing background appearance and inapparent object characteristics, which are visually hard to recognize.
Thus, the low-rank and sparsity regularizers are reasonable to be utilized to remove the trivial components in the image representation.
Consequently, the bilinear representation from MOMN is much more compact and generalized than previous methods, resulting in better fine-grained recognition performance.

\subsection{Comparison with state-of-the-art methods.}
In this section, we compare our MOMN with the state-of-the-art methods on five public datasets.
To evaluate the generalization of MOMN to different backbones, five backbones are employed, which are VGG-16, Densent-201, Resnet-50, Resnet-101, and Resnet-101\_32x8d \cite{Xie2016}.

We first evaluate the MOMN on three widely used benchmarks: CUB, Cars, and Aircraft, which are all widely used by FGVC methods.
Notably, since HBP~\cite{yu2018hierarchical} fuses the multi-level features to enhance their representation ability, we report their results of using a single $conv5\_3$ feature map for a fair comparison, denoted as FBP.
The results of DCL \cite{Chen_2019_CVPR} are reported without additional data augmentation strategies,~\emph{i.e.,} $14\times$ augmented data number of ours.
From Table~\ref{tab:fgvc_obj}, the proposed MOMN achieves the highest classification accuracy on all three datasets.
Compared with previous normalization-based methods, MOMN obtains about 1.0\% to 4.0\% improvements, which shows the superiority of our multi-objective normalization framework.
Especially, iSQRT (D=101) uses the strong SVM classifier, while our MOMN employs the weak Softmax classifier.

\begin{table}
\begin{center}
\caption{Evaluation of MOMN on Dogs and MPII datasets. The backbone of MOMN uses Resnet-101\_32x8d. } \label{tab:fgvc_dog_mpii}
\resizebox{0.9\columnwidth}{!}{
\begin{tabular}{lc|lc}
  \hline
   \multicolumn{2}{c|}{Dogs}&  \multicolumn{2}{c}{MPII} \\
  \hline
  Methods&ACC. (\%)&Methods&mAP (\%)\\
  \hline
  RAN\cite{Wang2017b}&83.1&RCNN\cite{Gkioxari2015}&27.7\\
  FCAN\cite{liu2016fully}&84.2&$\alpha$-pooling\cite{Simon2017}&30.8\\
  RACNN\cite{Fu2017}&87.3&HBP\cite{yu2018hierarchical}&32.4\\
  PC\cite{dubey2018pairwise}&83.8&Att.Pooling\cite{girdhar2017attentional}&30.6\\
  MAMC\cite{Sun2018}&87.3&iSQRT\cite{Li2018}&33.2\\
  \textbf{MOMN}&$\textbf{91.3}$&\textbf{MOMN}&$\textbf{34.3}$\\
  \hline
  
\end{tabular}}
\end{center}
\end{table}

Then, we further evaluate our MOMN on two additional datasets, which are Dogs and MPII, to demonstrate the excellent generalization of MOMN.
The results are reported in Table~\ref{tab:fgvc_dog_mpii}.
Impressively, MOMN outperforms the brand-new part-based method, on Dogs, by a large margin of 4\%. 
Also, in MPII, our MOMN wins the best performance among all existing methods.
Especially, MPII is a more challenging task for fine-grained action recognition, because it focuses on the interaction between human and background objects. 
The above comparisons verify that using MOMN with multi-objective normalization improves the discrimination of fine-grained representation, with superior generalization to different datasets.

Finally, we analyze the generalization ability of MOMN to different backbone architectures. 
As shown in Table~\ref{tab:fgvc_obj}, most existing methods are not well compatible with different backbone architectures. For example, KP \cite{cui2017kernel} and $\alpha$-pooling~\cite{Simon2017} both obtain worse performance on ResNet-50 than VGG-D backbones, and CBP and iSQRT cannot obtain further improvements after using deeper Densenet-201.
The reason may be that, as deep features are hard to train, these methods can not learn stabilized bilinear features.
In contrast, our MOMN shows great compatibility to all backbones, \emph{i.e.,} VGG-16, Densenet-201, and different Resnet variants.
Especially with Densenet-201, the results by MOMN are impressive.
Further, we use the Resnet-101\_32x8d as the backbone, and MOMN obtains the state-of-the-art performance.
The above comparison demonstrates that our MOMN has a better generalization ability to different backbones.

From the above evaluations, we summarize that: 1) MOMN is robust for fine-grained representation learning, and 2) MOMN has a better generalization ability for different backbones and datasets.


\section{Conclusion}
\label{sec:conclu}
Bilinear pooling suffers from unstable second-order information, redundant features, and over-fitting problems, which require different regularizers to process.
Existing methods cannot be adapted to multi-objective normalization with efficient implementation. 
Thus, in this paper, we propose a novel MOMN method that can normalize a bilinear representation, in terms of square-root, low-rank, and sparsity, simultaneously. 
In contrast to most methods considering only square-root, the extra low-rank and sparsity regularizers can eliminate the trivial components and avoid over-fitting, with negligible time addition.
To jointly optimize the above three non-smooth regularizers, an augmented Lagrange formula is designed with specific updating strategies based on gradient descent.
This allows different regularizers to be optimized alternately and iteratively.
Consequently, MOMN can not only normalize the bilinear representation in terms of various regularizers, but also be efficiently deployed in GPU platform with only matrix multiplication.
The evaluations on five benchmarks demonstrate the superior performance of MOMN in FGVC. 

In the future, more matrix regularizers, ~\emph{e.g.,} whitening and orthogonalization, will be explored under the proposed multi-objective normalization framework.
Further, MOMN will be extended to other applications, where the feature structures remain to be explored.


\ifCLASSOPTIONcaptionsoff
  \newpage
\fi



\bibliographystyle{IEEEtran}
\bibliography{ref}

\begin{thebibliography}{10}
\providecommand{\url}[1]{#1}
\csname url@samestyle\endcsname
\providecommand{\newblock}{\relax}
\providecommand{\bibinfo}[2]{#2}
\providecommand{\BIBentrySTDinterwordspacing}{\spaceskip=0pt\relax}
\providecommand{\BIBentryALTinterwordstretchfactor}{4}
\providecommand{\BIBentryALTinterwordspacing}{\spaceskip=\fontdimen2\font plus
\BIBentryALTinterwordstretchfactor\fontdimen3\font minus
  \fontdimen4\font\relax}
\providecommand{\BIBforeignlanguage}[2]{{%
\expandafter\ifx\csname l@#1\endcsname\relax
\typeout{** WARNING: IEEEtran.bst: No hyphenation pattern has been}%
\typeout{** loaded for the language `#1'. Using the pattern for}%
\typeout{** the default language instead.}%
\else
\language=\csname l@#1\endcsname
\fi
#2}}
\providecommand{\BIBdecl}{\relax}
\BIBdecl

\bibitem{10.1007/978-3-319-10590-1_54}
N.~Zhang, J.~Donahue, R.~Girshick, and T.~Darrell, ``Part-based r-cnns for
  fine-grained category detection,'' in \emph{Computer Vision -- ECCV 2014},
  D.~Fleet, T.~Pajdla, B.~Schiele, and T.~Tuytelaars, Eds.\hskip 1em plus 0.5em
  minus 0.4em\relax Cham: Springer International Publishing, 2014, pp.
  834--849.

\bibitem{Lin2015a}
D.~Lin, X.~Shen, C.~Lu, and J.~Jia, ``Deep lac: Deep localization, alignment
  and classification for fine-grained recognition,'' in \emph{Proceedings of
  the IEEE conference on computer vision and pattern recognition}, 2015, pp.
  1666--1674.

\bibitem{yao2016coarse}
H.~Yao, S.~Zhang, Y.~Zhang, J.~Li, and Q.~Tian, ``Coarse-to-fine description
  for fine-grained visual categorization,'' \emph{IEEE Transactions on Image
  Processing}, vol.~25, no.~10, pp. 4858--4872, 2016.

\bibitem{He2017}
X.~He and Y.~Peng, ``Weakly supervised learning of part selection model with
  spatial constraints for fine-grained image classification.'' in \emph{AAAI},
  2017, pp. 4075--4081.

\bibitem{ijcai2018-514}
F.~Zheng, X.~Miao, and H.~Huang, ``Fast vehicle identification in surveillance
  via ranked semantic sampling based embedding,'' in \emph{Proceedings of the
  Twenty-Seventh International Joint Conference on Artificial
  Intelligence}.\hskip 1em plus 0.5em minus 0.4em\relax International Joint
  Conferences on Artificial Intelligence Organization, 7 2018, pp. 3697--3703.

\bibitem{Wang2018}
Y.~Wang, V.~I. Morariu, and L.~S. Davis, ``Learning a discriminative filter
  bank within a cnn for fine-grained recognition,'' in \emph{Proceedings of the
  IEEE conference on computer vision and pattern recognition}, 2018, pp.
  4148--4157.

\bibitem{Lin2015}
T.-Y. Lin, A.~RoyChowdhury, and S.~Maji, ``Bilinear cnn models for fine-grained
  visual recognition,'' in \emph{Proceedings of the IEEE international
  conference on computer vision}, 2015, pp. 1449--1457.

\bibitem{sanchez2013image}
J.~S{\'a}nchez, F.~Perronnin, T.~Mensink, and J.~Verbeek, ``Image
  classification with the fisher vector: Theory and practice,''
  \emph{International journal of computer vision}, vol. 105, no.~3, pp.
  222--245, 2013.

\bibitem{wei2018grassmann}
X.~Wei, Y.~Zhang, Y.~Gong, J.~Zhang, and N.~Zheng, ``Grassmann pooling as
  compact homogeneous bilinear pooling for fine-grained visual
  classification,'' in \emph{Proceedings of the European Conference on Computer
  Vision (ECCV)}, 2018, pp. 355--370.

\bibitem{gou2018monet}
M.~Gou, F.~Xiong, O.~Camps, and M.~Sznaier, ``Monet: Moments embedding
  network,'' in \emph{Proceedings of the IEEE Conference on Computer Vision and
  Pattern Recognition}, 2018, pp. 3175--3183.

\bibitem{kong2017low}
S.~Kong and C.~Fowlkes, ``Low-rank bilinear pooling for fine-grained
  classification,'' in \emph{Proceedings of the IEEE conference on computer
  vision and pattern recognition}.\hskip 1em plus 0.5em minus 0.4em\relax IEEE,
  2017, pp. 7025--7034.

\bibitem{gao2016compact}
Y.~Gao, O.~Beijbom, N.~Zhang, and T.~Darrell, ``Compact bilinear pooling,'' in
  \emph{Proceedings of the IEEE conference on computer vision and pattern
  recognition}, 2016, pp. 317--326.

\bibitem{li2017second}
P.~Li, J.~Xie, Q.~Wang, and W.~Zuo, ``Is second-order information helpful for
  large-scale visual recognition,'' in \emph{Proceedings of the IEEE
  international conference on computer vision}, 2017, pp. 2070--2078.

\bibitem{Wang2017}
Q.~Wang, P.~Li, and L.~Zhang, ``G2denet: Global gaussian distribution embedding
  network and its application to visual recognition,'' in \emph{Proceedings of
  the IEEE conference on computer vision and pattern recognitionPR}, vol.~1,
  no.~2, 2017, p.~3.

\bibitem{lin2017improved}
T.-Y. Lin and S.~Maji, ``Improved bilinear pooling with cnns,'' \emph{BMVC},
  2017.

\bibitem{Li2018}
P.~Li, J.~Xie, Q.~Wang, and Z.~Gao, ``Towards faster training of global
  covariance pooling networks by iterative matrix square root normalization,''
  in \emph{Proceedings of the IEEE conference on computer vision and pattern
  recognition}, 2018.

\bibitem{cui2017kernel}
Y.~Cui, F.~Zhou, J.~Wang, X.~Liu, Y.~Lin, and S.~Belongie, ``Kernel pooling for
  convolutional neural networks,'' in \emph{Proceedings of the IEEE conference
  on computer vision and pattern recognition}, vol.~1, no.~2, 2017, p.~7.

\bibitem{cai2017higher}
S.~Cai, W.~Zuo, and L.~Zhang, ``Higher-order integration of hierarchical
  convolutional activations for fine-grained visual categorization,'' in
  \emph{Proceedings of the IEEE International Conference on Computer Vision},
  2017, pp. 511--520.

\bibitem{ionescu2015matrix}
C.~Ionescu, O.~Vantzos, and C.~Sminchisescu, ``Matrix backpropagation for deep
  networks with structured layers,'' in \emph{Proceedings of the IEEE
  International Conference on Computer Vision}, 2015, pp. 2965--2973.

\bibitem{EnginWZL18}
M.~Engin, L.~Wang, L.~Zhou, and X.~Liu, ``Deepkspd: Learning
  kernel-matrix-based {SPD} representation for fine-grained image
  recognition,'' in \emph{Proceedings of the European Conference on Computer
  Vision}, 2018, pp. 629--645.

\bibitem{Higham1997}
N.~J. Higham, ``Stable iterations for the matrix square root,'' \emph{Numerical
  Algorithms}, vol.~15, no.~2, pp. 227--242, 1997.

\bibitem{Higham2009}
N.~Higham, ``Functions of matrices: Theory and computation, siam, philadelphia,
  2008,'' \emph{Johann Radon Institute (RICAM)}, 2009.

\bibitem{roberts1980linear}
J.~D. Roberts, ``Linear model reduction and solution of the algebraic riccati
  equation by use of the sign function,'' \emph{International Journal of
  Control}, vol.~32, no.~4, pp. 677--687, 1980.

\bibitem{boyd2011distributed}
S.~Boyd, N.~Parikh, E.~Chu, B.~Peleato, J.~Eckstein \emph{et~al.},
  ``Distributed optimization and statistical learning via the alternating
  direction method of multipliers,'' \emph{Foundations and Trends in Machine
  learning}, vol.~3, no.~1, pp. 1--122, 2011.

\bibitem{gao2013learning}
S.~Gao, I.~W.-H. Tsang, and Y.~Ma, ``Learning category-specific dictionary and
  shared dictionary for fine-grained image categorization,'' \emph{IEEE
  Transactions on Image Processing}, vol.~23, no.~2, pp. 623--634, 2013.

\bibitem{zhang2015detecting}
L.~Zhang, Y.~Yang, M.~Wang, R.~Hong, L.~Nie, and X.~Li, ``Detecting densely
  distributed graph patterns for fine-grained image categorization,''
  \emph{IEEE Transactions on Image Processing}, vol.~25, no.~2, pp. 553--565,
  2015.

\bibitem{liu2016hierarchical}
A.-A. Liu, Y.-T. Su, W.-Z. Nie, and M.~Kankanhalli, ``Hierarchical clustering
  multi-task learning for joint human action grouping and recognition,''
  \emph{IEEE transactions on pattern analysis and machine intelligence},
  vol.~39, no.~1, pp. 102--114, 2016.

\bibitem{zhang2012attribute}
H.~Zhang, Z.-J. Zha, S.~Yan, J.~Bian, and T.-S. Chua, ``Attribute feedback,''
  in \emph{Proceedings of the 20th ACM international conference on Multimedia},
  2012, pp. 79--88.

\bibitem{zhang2014robust}
H.~Zhang, Z.-J. Zha, Y.~Yang, S.~Yan, and T.-S. Chua, ``Robust (semi)
  nonnegative graph embedding,'' \emph{IEEE transactions on image processing},
  vol.~23, no.~7, pp. 2996--3012, 2014.

\bibitem{Simon2017}
M.~Simon, Y.~Gao, T.~Darrell, J.~Denzler, and E.~Rodner, ``Generalized
  orderless pooling performs implicit salient matching,'' in \emph{Proceedings
  of the IEEE international conference on computer vision}, 2017.

\bibitem{xu2019multi}
N.~Xu, H.~Zhang, A.-A. Liu, W.~Nie, Y.~Su, J.~Nie, and Y.~Zhang, ``Multi-level
  policy and reward-based deep reinforcement learning framework for image
  captioning,'' \emph{IEEE Transactions on Multimedia}, 2019.

\bibitem{7226712}
H.~{Yao}, S.~{Zhang}, F.~{Xie}, Y.~{Zhang}, D.~{Zhang}, Y.~{Su}, and Q.~{Tian},
  ``Orientational spatial part modeling for fine-grained visual
  categorization,'' in \emph{2015 IEEE International Conference on Mobile
  Services}, June 2015, pp. 360--367.

\bibitem{reed2016learning}
S.~Reed, Z.~Akata, H.~Lee, and B.~Schiele, ``Learning deep representations of
  fine-grained visual descriptions,'' in \emph{Proceedings of the IEEE
  Conference on Computer Vision and Pattern Recognition}, 2016, pp. 49--58.

\bibitem{yao2017dsp}
H.~Yao, D.~Zhang, J.~Li, J.~Zhou, S.~Zhang, and Y.~Zhang, ``Dsp: Discriminative
  spatial part modeling for fine-grained visual categorization,'' \emph{Image
  and Vision Computing}, vol.~63, pp. 24--37, 2017.

\bibitem{Liu2017}
X.~Liu, J.~Wang, S.~Wen, E.~Ding, and Y.~Lin, ``Localizing by describing:
  Attribute-guided attention localization for fine-grained recognition.'' in
  \emph{AAAI}, 2017, pp. 4190--4196.

\bibitem{Wei2018}
L.~Wei, X.~Wang, A.~Wu, R.~Zhou, and C.~Zhu, ``Robust subspace segmentation by
  self-representation constrained low-rank representation,'' \emph{Neural
  Processing Letters}, pp. 1--21, 2018.

\bibitem{zhang2019learning}
L.~Zhang, S.~Huang, W.~Liu, and D.~Tao, ``Learning a mixture of
  granularity-specific experts for fine-grained categorization,'' in
  \emph{Proceedings of the IEEE International Conference on Computer Vision},
  2019, pp. 8331--8340.

\bibitem{ding2019selective}
Y.~Ding, Y.~Zhou, Y.~Zhu, Q.~Ye, and J.~Jiao, ``Selective sparse sampling for
  fine-grained image recognition,'' in \emph{Proceedings of the IEEE
  International Conference on Computer Vision}, 2019, pp. 6599--6608.

\bibitem{Jaderberg2015}
M.~Jaderberg, K.~Simonyan, A.~Zisserman \emph{et~al.}, ``Spatial transformer
  networks,'' in \emph{Advances in Neural Information Processing Systems},
  2015, pp. 2017--2025.

\bibitem{Simon2015}
M.~Simon and E.~Rodner, ``Neural activation constellations: Unsupervised part
  model discovery with convolutional networks,'' in \emph{Proceedings of the
  IEEE international conference on computer vision}, 2015, pp. 1143--1151.

\bibitem{krause2015fine}
J.~Krause, H.~Jin, J.~Yang, and L.~Fei-Fei, ``Fine-grained recognition without
  part annotations,'' in \emph{Proceedings of the IEEE Conference on Computer
  Vision and Pattern Recognition}, 2015, pp. 5546--5555.

\bibitem{Fu2017}
J.~Fu, H.~Zheng, and T.~Mei, ``Look closer to see better: Recurrent attention
  convolutional neural network for fine-grained image recognition,'' in
  \emph{Proceedings of the IEEE conference on computer vision and pattern
  recognition}, 2017, pp. 4438--4446.

\bibitem{yao2017autobd}
H.~Yao, S.~Zhang, C.~Yan, Y.~Zhang, J.~Li, and Q.~Tian, ``Autobd: Automated
  bi-level description for scalable fine-grained visual categorization,''
  \emph{IEEE Transactions on Image Processing}, vol.~27, no.~1, pp. 10--23,
  2017.

\bibitem{tuzel2006region}
O.~Tuzel, F.~Porikli, and P.~Meer, ``Region covariance: A fast descriptor for
  detection and classification,'' in \emph{European conference on computer
  vision}.\hskip 1em plus 0.5em minus 0.4em\relax Springer, 2006, pp. 589--600.

\bibitem{tuzel2007human}
O.~Tuzel, F.~Porikli, P.~Meer \emph{et~al.}, ``Human detection via
  classification on riemannian manifolds.'' in \emph{CVPR}, vol.~1, no.~2,
  2007, p.~4.

\bibitem{min2019adaptive}
S.~Min, H.~Xie, Y.~Tian, H.~Yao, and Y.~Zhang, ``Adaptive bilinear pooling for
  fine-grained representation learning,'' in \emph{Proceedings of the ACM
  Multimedia Asia on ZZZ}, 2019, pp. 1--6.

\bibitem{charikar2004finding}
M.~Charikar, K.~Chen, and M.~Farach-Colton, ``Finding frequent items in data
  streams,'' \emph{Theoretical Computer Science}, vol. 312, no.~1, pp. 3--15,
  2004.

\bibitem{Pham2013}
N.~Pham and R.~Pagh, ``Fast and scalable polynomial kernels via explicit
  feature maps,'' in \emph{Proceedings of the 19th ACM SIGKDD international
  conference on Knowledge discovery and data mining}.\hskip 1em plus 0.5em
  minus 0.4em\relax ACM, 2013, pp. 239--247.

\bibitem{yu2018hierarchical}
C.~Yu, X.~Zhao, Q.~Zheng, P.~Zhang, and X.~You, ``Hierarchical bilinear pooling
  for fine-grained visual recognition,'' in \emph{Proceedings of the European
  Conference on Computer Vision}.\hskip 1em plus 0.5em minus 0.4em\relax
  Springer, 2018, pp. 595--610.

\bibitem{book}
A.~Gil, J.~Segura, and N.~Temme, \emph{Numerical Methods for Special
  Functions}, 01 2007.

\bibitem{Zhang2013}
T.~Zhang, B.~Ghanem, S.~Liu, C.~Xu, and N.~Ahuja, ``Low-rank sparse coding for
  image classification,'' in \emph{Proceedings of the IEEE international
  conference on computer visionV}, 2013, pp. 281--288.

\bibitem{fan2018virtual}
Z.~Fan, D.~Zhang, X.~Wang, Q.~Zhu, and Y.~Wang, ``Virtual dictionary based
  kernel sparse representation for face recognition,'' \emph{Pattern
  Recognition}, vol.~76, pp. 1--13, 2018.

\bibitem{xu2017multi}
J.~Xu, L.~Zhang, D.~Zhang, and X.~Feng, ``Multi-channel weighted nuclear norm
  minimization for real color image denoising,'' in \emph{Proceedings of the
  IEEE International Conference on Computer Vision}, 2017, pp. 1096--1104.

\bibitem{wei2018robust}
L.~Wei, X.~Wang, A.~Wu, R.~Zhou, and C.~Zhu, ``Robust subspace segmentation by
  self-representation constrained low-rank representation,'' \emph{Neural
  Processing Letters}, vol.~48, no.~3, pp. 1671--1691, 2018.

\bibitem{yang2018lr}
W.~Yang, X.~Shang, K.~Chen, and S.~Sun, ``Lr 2-sr: Laplacian regularized
  low-rank sparse representation for single image super-resolution,'' in
  \emph{2018 IEEE Fourth International Conference on Multimedia Big Data
  (BigMM)}.\hskip 1em plus 0.5em minus 0.4em\relax IEEE, 2018, pp. 1--4.

\bibitem{sener2018multi}
O.~Sener and V.~Koltun, ``Multi-task learning as multi-objective
  optimization,'' in \emph{Advances in Neural Information Processing Systems},
  2018, pp. 527--538.

\bibitem{miettinen2012nonlinear}
K.~Miettinen, \emph{Nonlinear multiobjective optimization}.\hskip 1em plus
  0.5em minus 0.4em\relax Springer Science \& Business Media, 2012, vol.~12.

\bibitem{ehrgott2005multicriteria}
M.~Ehrgott, \emph{Multicriteria optimization}.\hskip 1em plus 0.5em minus
  0.4em\relax Springer Science \& Business Media, 2005, vol. 491.

\bibitem{fliege2000steepest}
J.~Fliege and B.~F. Svaiter, ``Steepest descent methods for multicriteria
  optimization,'' \emph{Mathematical Methods of Operations Research}, vol.~51,
  no.~3, pp. 479--494, 2000.

\bibitem{schaffler2002stochastic}
S.~Sch{\"a}ffler, R.~Schultz, and K.~Weinzierl, ``Stochastic method for the
  solution of unconstrained vector optimization problems,'' \emph{Journal of
  Optimization Theory and Applications}, vol. 114, no.~1, pp. 209--222, 2002.

\bibitem{desideri2012multiple}
J.-A. D{\'e}sid{\'e}ri, ``Multiple-gradient descent algorithm (mgda) for
  multiobjective optimization,'' \emph{Comptes Rendus Mathematique}, vol. 350,
  no. 5-6, pp. 313--318, 2012.

\bibitem{peitz2018gradient}
S.~Peitz and M.~Dellnitz, ``Gradient-based multiobjective optimization with
  uncertainties,'' in \emph{NEO 2016}.\hskip 1em plus 0.5em minus 0.4em\relax
  Springer, 2018, pp. 159--182.

\bibitem{bertsekas1997nonlinear}
D.~P. Bertsekas, ``Nonlinear programming,'' \emph{Journal of the Operational
  Research Society}, vol.~48, no.~3, pp. 334--334, 1997.

\bibitem{you2014diverse}
X.~You, R.~Wang, and D.~Tao, ``Diverse expected gradient active learning for
  relative attributes,'' \emph{IEEE Transactions on Image Processing}, vol.~23,
  no.~7, pp. 3203--3217, 2014.

\bibitem{poirion2017descent}
F.~Poirion, Q.~Mercier, and J.-A. D{\'e}sid{\'e}ri, ``Descent algorithm for
  nonsmooth stochastic multiobjective optimization,'' \emph{Computational
  Optimization and Applications}, vol.~68, no.~2, pp. 317--331, 2017.

\bibitem{huang2016part}
S.~Huang, Z.~Xu, D.~Tao, and Y.~Zhang, ``Part-stacked cnn for fine-grained
  visual categorization,'' in \emph{Proceedings of the IEEE conference on
  computer vision and pattern recognition}, 2016, pp. 1173--1182.

\bibitem{8499082}
W.~Yang, X.~Shang, K.~Chen, and S.~Sun, ``Lr2-sr: Laplacian regularized
  low-rank sparse representation for single image super-resolution,'' in
  \emph{BigMM}, Sep. 2018, pp. 1--4.

\bibitem{Hu2018}
J.~Hu, L.~Shen, and G.~Sun, ``Squeeze-and-excitation networks,'' in
  \emph{Proceedings of the IEEE conference on computer vision and pattern
  recognition}, 2018, pp. 7132--7141.

\bibitem{wah2011caltech}
C.~Wah, S.~Branson, P.~Welinder, P.~Perona, and S.~Belongie, ``The caltech-ucsd
  birds-200-2011 dataset,'' 2011.

\bibitem{Krause2013}
J.~Krause, M.~Stark, J.~Deng, and L.~Fei-Fei, ``3d object representations for
  fine-grained categorization,'' in \emph{Proceedings of the IEEE international
  conference on computer vision Workshops}, 2013, pp. 554--561.

\bibitem{Maji2013}
S.~Maji, E.~Rahtu, J.~Kannala, M.~Blaschko, and A.~Vedaldi, ``Fine-grained
  visual classification of aircraft,'' \emph{HAL-INRIA}, p.~5, 2013.

\bibitem{khosla2011novel}
A.~Khosla, N.~Jayadevaprakash, B.~Yao, and F.-F. Li, ``Novel dataset for
  fine-grained image categorization: Stanford dogs,'' in \emph{Proc. CVPR
  Workshop on Fine-Grained Visual Categorization (FGVC)}, vol.~2, no.~1, 2011.

\bibitem{Pishchulin2014}
L.~Pishchulin, M.~Andriluka, and B.~Schiele, ``Fine-grained activity
  recognition with holistic and pose based features,'' in \emph{GCPR}.\hskip
  1em plus 0.5em minus 0.4em\relax Springer, 2014, pp. 678--689.

\bibitem{Gkioxari2015}
G.~Gkioxari, R.~Girshick, and J.~Malik, ``Contextual action recognition with r*
  cnn,'' in \emph{Proceedings of the IEEE international conference on computer
  vision}, 2015, pp. 1080--1088.

\bibitem{girdhar2017attentional}
R.~Girdhar and D.~Ramanan, ``Attentional pooling for action recognition,'' in
  \emph{Advances in Neural Information Processing Systems}, 2017, pp. 34--45.

\bibitem{he2016deep}
K.~He, X.~Zhang, S.~Ren, and J.~Sun, ``Deep residual learning for image
  recognition,'' in \emph{Proceedings of the IEEE conference on computer vision
  and pattern recognition}, 2016, pp. 770--778.

\bibitem{dubey2018pairwise}
A.~Dubey, O.~Gupta, P.~Guo, R.~Raskar, R.~Farrell, and N.~Naik, ``Pairwise
  confusion for fine-grained visual classification,'' in \emph{Proceedings of
  the European Conference on Computer Vision}, 2018, pp. 70--86.

\bibitem{Chen_2019_CVPR}
Y.~Chen, Y.~Bai, W.~Zhang, and T.~Mei, ``Destruction and construction learning
  for fine-grained image recognition,'' in \emph{The IEEE Conference on
  Computer Vision and Pattern Recognition}, June 2019.

\bibitem{Xie2016}
S.~Xie, R.~Girshick, P.~Dollár, Z.~Tu, and K.~He, ``Aggregated residual
  transformations for deep neural networks,'' \emph{arXiv preprint
  arXiv:1611.05431}, 2016.

\bibitem{Wang2017b}
F.~Wang, M.~Jiang, C.~Qian, S.~Yang, C.~Li, H.~Zhang, X.~Wang, and X.~Tang,
  ``Residual attention network for image classification,'' in \emph{Proceedings
  of the IEEE Conference on Computer Vision and Pattern Recognition}, 2017, pp.
  3156--3164.

\bibitem{liu2016fully}
X.~Liu, T.~Xia, J.~Wang, Y.~Yang, F.~Zhou, and Y.~Lin, ``Fully convolutional
  attention networks for fine-grained recognition,'' \emph{arXiv preprint
  arXiv:1603.06765}, 2016.

\bibitem{Sun2018}
M.~Sun, Y.~Yuan, F.~Zhou, and E.~Ding, ``Multi-attention multi-class constraint
  for fine-grained image recognition,'' in \emph{Proceedings of the European
  Conference on Computer Vision}, 2018, pp. 805--821.

\end{thebibliography}
%
\end{document}